\documentclass[fleqn,10pt]{wlscirep}
\usepackage[utf8]{inputenc}
\usepackage[T1]{fontenc}
\usepackage{bm}
\usepackage{amssymb}
\usepackage[utf8]{inputenc} 
\usepackage[T1]{fontenc}    
\usepackage{hyperref}       
\usepackage{url}            
\usepackage{booktabs}       
\usepackage{amsfonts}       
\usepackage{nicefrac}       
\usepackage{microtype}      
\usepackage{xcolor}         
\usepackage{epsfig}
\usepackage{graphicx}
\usepackage{amsmath}
\usepackage{graphicx}
\usepackage{amssymb}
\usepackage{booktabs}
\usepackage{graphicx}
\usepackage{float}
\usepackage{stfloats}
\usepackage{epstopdf}
\usepackage{algorithm,algorithmic}
\usepackage{multirow}
\usepackage{bbm}
\usepackage{framed}
\usepackage{amsthm}
\usepackage{adjustbox}
\usepackage{color}
\usepackage{lineno}
\usepackage{url}
\usepackage{cleveref}
\usepackage{diagbox}
\usepackage{colortbl}
\usepackage{arydshln}
\usepackage{multirow}
\usepackage{multicol}
\usepackage{makecell}
\usepackage{upgreek}
\usepackage{gensymb}
\usepackage{lineno}
\usepackage{makecell}
\usepackage{adjustbox}
\usepackage{tabularx}
\usepackage[utf8]{inputenc}
\newcommand{\blue}[1]{\textcolor{black}{#1}}

\title{Physics-aware Machine Learning Revolutionizes Scientific Paradigm for Machine Learning and Process-based Hydrology}

\author[1]{Qingsong Xu}
\author[3]{Yilei Shi}
\author[1,6]{Jonathan Bamber}
\author[4]{Ye Tuo}
\author[5]{Ralf Ludwig}
\author[1,2]{Xiao Xiang Zhu}
\affil[1]{Data Science in Earth Observation, Technical University of Munich, Munich, Germany}
\affil[2]{Munich Center for Machine Learning,  Munich, Germany}
\affil[3]{School of Engineering and Design, Technical University of Munich, Munich, Germany}
\affil[4]{Hydrology and River Basin Management, Technical University of Munich, Munich, Germany}
\affil[5]{Department of Geography, Ludwig-Maximilians-University, Munich, Germany}
\affil[6]{School of Geographical Sciences, University of Bristol, UK}

\begin{abstract}
Accurate geoscientific process understanding and water cycle prediction are crucial for addressing scientific and societal challenges associated with the management of water resources. Existing reviews predominantly concentrate on the development of machine learning (ML) in this field, yet there is a clear distinction between process-based hydrology and ML as separate paradigms. Here, we introduce physics-aware ML as a transformative approach to overcome the perceived barrier and revolutionize both fields. Specifically, we present a comprehensive review of the physics-aware ML methods, building a structured community (PaML) of existing methodologies that integrate prior physical knowledge or physics-based modeling into ML. We systematically analyze these PaML methodologies with respect to four aspects: physical data-guided ML, physics-informed ML, physics-embedded ML, and physics-aware hybrid learning. PaML facilitates ML-aided hypotheses, accelerating insights from big data and fostering scientific discoveries. We first conduct a systematic review of hydrology in PaML, including rainfall-runoff hydrological processes and hydrodynamic processes, and highlight the most promising and challenging directions for different objectives and PaML methods. Finally, a new PaML-based hydrology platform, termed HydroPML, is released as a foundation for hydrological applications. HydroPML enhances the explainability and causality of ML and lays the groundwork for the digital water cycle's realization. The HydroPML platform is publicly available at \textcolor{blue}{https://hydropml.github.io/}.
\end{abstract}
\begin{document}

\flushbottom
\maketitle

\thispagestyle{empty}

\section{Introduction}
Numerous scientific and societal challenges associated with understanding and preparing for environmental change rest upon our ability to understand and predict water cycle changes~\cite{bloschl2019twenty}. Process-based hydrological models play a crucial role in understanding and managing the Earth's water resources~\cite{devia2015review} and planning for water security and managing extremes such as floods and droughts~\cite{brunner2021challenges}. 
Many hydrological processes  can be described as complex physical dynamics spanning various spatial and temporal scales, such as hydrodynamic processes and rainfall-runoff processes.  Physical methods have been highly effective in elucidating and forecasting the state changes in a hydrological process~\cite{dingman2015physical}. However, process-based hydrology continues to present important challenges~\cite{bloschl2019twenty}. (1) A subset of process-based hydrological methods requires significant computational resources and expertise. For example, a global flood solver requires dramatic computational resources based on a traditional finite difference solver~\cite{zhou2021toward}.
(2) Process-based hydrological models encounter limitations due to existing knowledge gaps. For example,  the two-way feedback between humans and water systems is physically agnostic, but it is crucial for the water cycles~\cite{wang2021predicting, meempatta2019reviewing}.  A new paradigm is needed to reduce knowledge gaps and explore the unknowns.
(3) Process-based hydrology models frequently struggle to rapidly exploit the information in big data.  For example,  abundant remote sensing observations and hydrological measurements can be utilized for parameter calibration by reducing the differences between model predictions and these big data. However, 
most traditional process-based hydrological models rely on iterative parameter calibration, which introduces a significant level of uncertainty. Additionally, the iterative process incurs high computational costs~\cite{fatichi2016overview}.

Machine learning (ML) provides efficient alternatives for learning dynamic geophysical phenomena from massive datasets~\cite {jiang2020improving}. Recent works have shown that ML can generate realistic short-term hydrology predictions~\cite{kratzert2018rainfall} and significantly speed up the simulation of hydrodynamic processes~\cite{vinuesa2022enhancing}. Despite the tremendous progress, ML is purely data-driven by nature, which presents many limitations~\cite{wang2021physics, shen2023differentiable}. (1) The nonlinear and chaotic nature of process-based hydrology poses significant challenges for existing data-driven frameworks, which still adhere to the fundamental principles of statistical inference. (2) ML models often struggle with maintaining physical consistency. In the absence of explicit constraints, data-driven models can produce forecasts that violate the physical laws of the hydrological process, leading to implausible outcomes or plausible, yet unexplainable results. (3) Purely data-driven ML models struggle with generalization, as they face challenges in predicting untrained variables and adapting to unseen scenarios with different distributions. This is especially pronounced in process-based systems, where nonlinearity and changing system parameters contribute to distribution shifts. (4) Purely data-driven models lack interpretability and causality, which compromises the reliability of their projections when circumstances change.

Both ML alone and purely physics-based approaches are inadequate for effectively learning complex dynamic processes, such as hydrodynamic and rainfall-runoff processes. Thus, there is an increasing demand to integrate traditional physics-based approaches with ML models, combining the strengths of both approaches. While existing research on physics-aware ML ~\cite{wang2021physics, lawal2022physics, khandelwal2020physics, cai2021physics} is extensive, it is not exhaustive or sufficiently comprehensive to encompass a broad range of research domains related to dynamic processes, particularly hydrological processes.

To our knowledge, there is no systematic summary and analysis of physics-aware ML  in general. The ever-growing quantity of physics-aware ML methods makes
it difficult to find the proper one for a specific problem of dynamic processes. For example, Lawal et al.~\cite{lawal2022physics} study and assess physics-informed neural networks (PINNs) from various researchers’ perspectives, categorizing newly introduced PINN methodologies into extended PINNs, hybrid PINNs, and techniques for minimizing loss. A review of the three neural network frameworks~\cite{faroughi2022physics} (i.e., physics-guided neural networks, PINNs, and physics-encoded neural networks) is presented and analyzed. In addition, Goswami et al.~\cite{goswami2022physics} present a review of deep neural operator networks and appropriate extensions with physics-informed deep neural operators. An  overview of existing physics-guided deep learning (DL)~\cite{wang2021physics} is provided, and existing physics-guided DL approaches are categorized into physics-guided loss function, physics-guided architecture design, hybrid physics-DL models, and invariant and equivariant DL models. In contrast, Willard et al.~\cite{willard2020integrating} categorize existing methodologies into physics-guided ML models and hybrid physics-ML frameworks.  However, at present, the information about physics-aware ML in these existing reviews is still limited, and these classification systems are also confusing.
Beyond summarization and categorization, insightful analysis of the benefits and limitations of existing physics-aware ML methods can help researchers understand the current research state and trends of the scientific ML community. Thus, systematic analysis of physics-aware ML is crucial to the
future development of physical dynamics.

There is also no review of physics-aware ML specific to process-based hydrology. Existing reviews are all related to ML in the field of hydrology, or conceptual reviews of physics-aware ML in geoscience, engineering and environmental systems, or Earth system science~\cite{camps2021deep}. For instance, an introductory review of ML for hydrologic sciences~\cite{xu2021machine} is intended for readers
new to the field of ML. Sit et al.~\cite{sit2020comprehensive} offer an extensive overview of DL methods applied in the water industry, covering tasks such as generation, enhancement, prediction,  and classification. In addition,   Reichstein et al.~\cite{reichstein2019deep} review the development of ML in the geoscientific context, highlighting DL's potential to overcome many of the limitations of applying ML. The challenging approaches to combining
ML with physical modeling are laid out. Furthermore, differentiable modeling~\cite{shen2023differentiable} that connects varying amounts of prior knowledge to neural networks (NNs) and trains them together is proposed to offer better interpretability, generalizability, and extrapolation capability for advanced geosciences. 
However, many hydrological processes can be described as dynamic systems of physical equations, such as shallow water equations in hydrodynamic processes, and mass conservation equations in rainfall-runoff processes. Physics-aware ML is important for hydrology in that it integrates physical principles and domain expertise, and enables accurate predictions and extrapolation in complex hydrological systems. Thus, building a review of physics-aware ML in process-based hydrology is urgently needed for physical hydrology.

\begin{figure*}[!tph]
	\centering
	{\includegraphics[width = 1.0\textwidth]{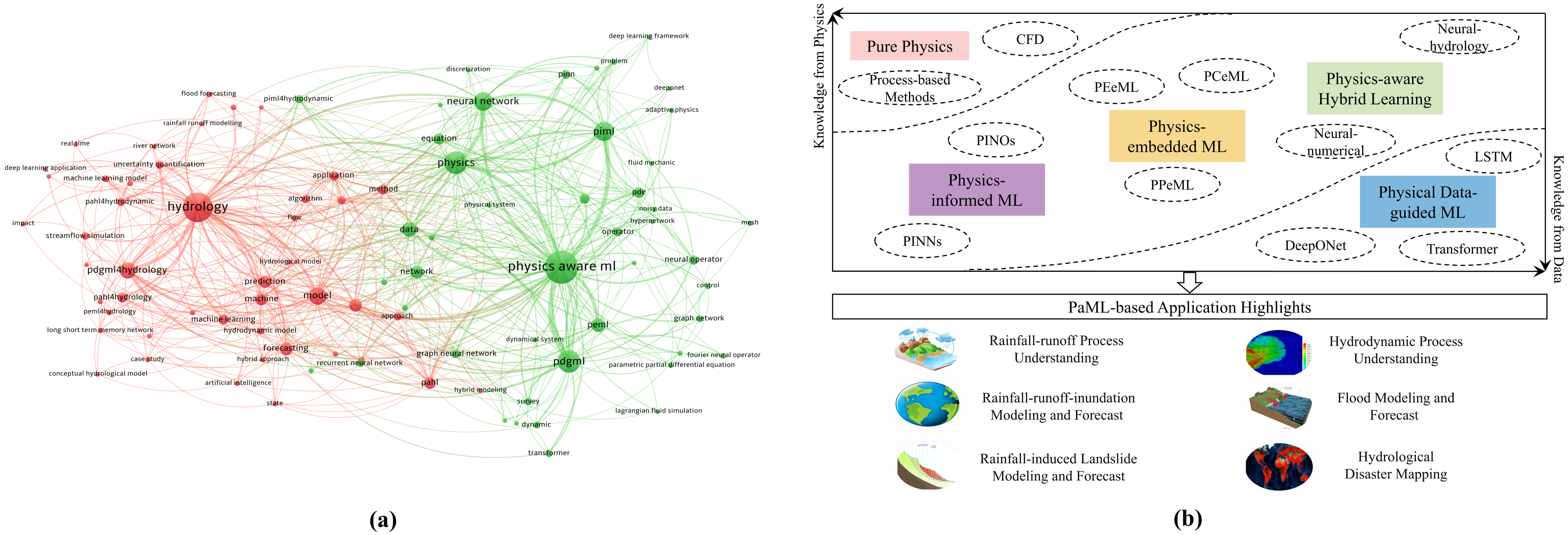}}
	\vspace{-2mm}
	\caption{(a) Examine the knowledge gap between physics-aware ML and hydrology in physics-aware ML from the keyword viewpoint.  (b) Conceptual framework of physics-aware ML (PaML) and PaML-based hydrological application highlights.  PaML includes physical data-guided ML, physics-informed ML, physics-embedded ML, and physics-aware hybrid learning.}
	\label{fig:00}
	\vspace{-2mm}
\end{figure*}

There is a significant  knowledge gap between physics-aware ML and process-based hydrology. Through extensive literature analysis (Fig.~\ref{fig:00}(a)), physics-aware ML is developing rapidly in the field of ML through different combinations of physics and ML. For example, partial differential equation (PDE) solutions are represented as NNs by including the square of the PDE residual in the loss function, which was developed in 1998~\cite{lagaris1998artificial}. In 2019, this approach was refined further and called PINNs~\cite{raissi2019physics}, initiating a flurry of follow-up work~\cite{karniadakis2021physics}. In 2022,
 vanilla PINNs began to be used in the field of hydrodynamics (flood) processes~\cite{mahesh2022physics, bihlo2022physics, bojovic2022physics}. In the ML field, in order to improve the interpretability, causality, and generalizability of ML, and rapidly capture accurate dynamic processes, dynamic physics has also been refined into  physical equations, physical properties, and different physical conditions. These fine-grained physical laws are embedded in the frameworks or modules of ML in different ways for different dynamic problems, employing end-to-end training techniques. However, the current integration of physics in hydrology into ML typically follows a simple data-driven or hybrid learning approach~\cite{slater2023hybrid}, which can lead to increased computational burden and potential biases in hydrological models. Thus, further investigation and analysis of physics-aware ML and hydrology in physics-aware ML are imperative to reduce and elucidate the knowledge gap between physics-aware ML and process-based hydrology.

 Bearing these concerns in mind, we make an exhaustive and comprehensive review of the physics-aware ML methods.  A structured community of existing methodologies that integrates prior physical knowledge or physics-based modeling into ML is built, and called PaML. As shown in Fig.~\ref{fig:00}(b), PaML approaches are categorized into four groups based on the way physics and ML are combined: (1) physical data-guided ML: a supervised DL model that statistically learns the   known or unknown physics of a desired phenomenon by extracting features or   attributes from raw training data; (2) physics-informed ML: a widely adopted method involves training models using supplementary information derived from enforcing physical constraints (for example, designing loss functions   (regularization)); (3) physics-embedded ML: embedding physics in the model frameworks or   modules in an end-to-end manner; and (4) physics-aware hybrid learning: directly combining pure physics-based models, such as   numerical methods  and hydrology models, with ML   models. A
systematic analysis of each of these different methods  is provided. 
Then, we conduct a systematic review of process-based hydrology in physics-aware ML, including hydrodynamic processes and rainfall-runoff hydrological processes. Next, PaML-based hydrodynamic modeling and PaML-based rainfall-runoff hydrological forecasts are classified by different objectives and PaML methods. 
We further highlight the most promising and challenging directions for different application categories, aiming to identify opportunities for advancement and address existing hurdles in the field. Finally, in order to reduce the knowledge gap between physics-aware ML and process-based hydrology,  a new
PaML-based hydrology platform, termed HydroPML, is proposed as a foundation
for applications based on hydrological processes. For example, we apply HydroPML to real-time flood forecasts.

The main contributions of this work are summarized below.
\begin{itemize}
\item We conduct an extensive and comprehensive review of physics-aware ML methods, which serve as a transformative approach to bridge process-based hydrology and ML, ultimately facilitating a paradigm shift in both fields.  These  methods are summarized and categorized into physical data-guided ML, physics-informed ML, physics-embedded ML, and physics-aware hybrid learning. 
\item Systematic analyses of four
aspects of these PaML methodologies are undertaken, to provide insights and
ideas for research within the PaML community.
\item We systematically analyze PaML-based hydrological processes, specifically focusing on hydrodynamic processes and rainfall-runoff hydrological processes. We categorize PaML-based hydrodynamic modeling and PaML-based rainfall-runoff hydrological forecasts based on different objectives and PaML methods.  Additionally, we identify the most promising directions and challenges in various application categories, aiming to advance the field and overcome existing obstacles.
\item We release an open platform, termed HydroPML, as a foundation for applications based on hydrological processes. HydroPML bridges the gap between PaML and process-based hydrology, offering a range of hydrology applications (as exemplified in Fig.~\ref{fig:00}(b)), including but not limited to rainfall-runoff-inundation modeling and forecasting, real-time flood modeling and forecasting, and cutting-edge PaML methods to enhance water security and fostering resilient water management.
\end{itemize}

The rest of this paper is organized as follows.  Section~\ref{section2} reviews state-of-the-art physics-aware ML. Section~\ref{section3} presents process-based hydrology in physics-aware ML (HydroPML), and its application highlights. Section~\ref{section4} concludes the paper.

\section{State-of-the-art Physics-aware Machine Learning \label{section2}}

\begin{figure}[!tph]
        \vspace{-4mm}
	\centering
	{\includegraphics[width = 0.8\textwidth]{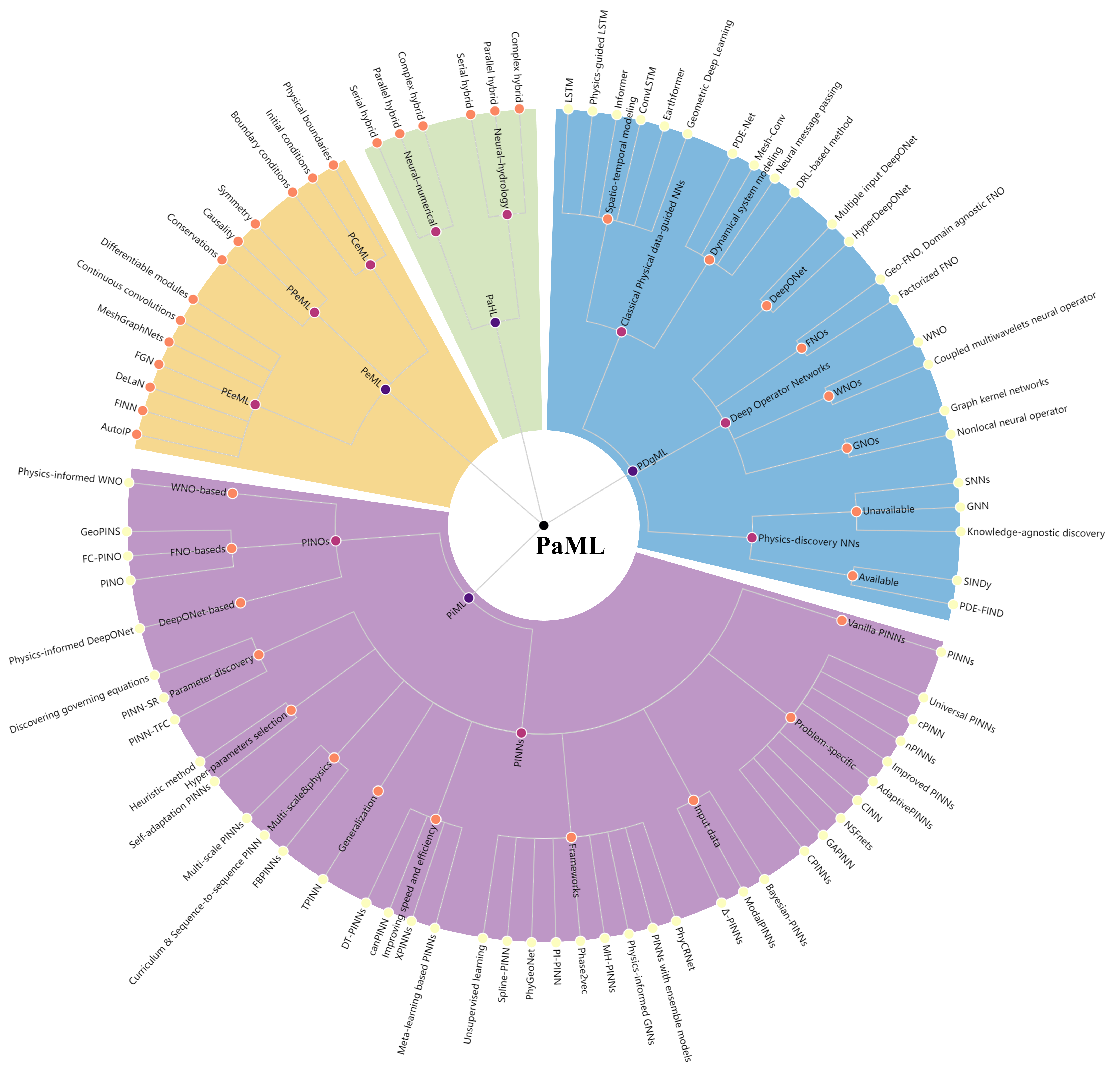}}
	\caption{The proposed physics-aware machine learning (PaML) community.}
	\label{fig:3}
\end{figure}
PaML aims to integrate the strengths of physics-based modeling and state-of-the-art ML models to effectively tackle scientific challenges. As shown in Fig.~\ref{fig:3},   a structured community of existing PaML methodologies that integrate prior physical knowledge or physics-based modeling into ML is built. We categorize PaML approaches into four groups based on the way physics and ML are combined, including physical data-guided ML (PDgML), physics-informed ML (PiML), physics-embedded ML (PeML), and physics-aware hybrid learning (PaHL). \blue{These four methods in the PaML community, including their corresponding benefits and drawbacks for scientific problems, are summarized in Table~\ref{tab4}.}  Each is discussed in detail below.
\begin{table}[!htp]
	\caption{\blue{Four approaches to physics-aware machine learning.}}
	\centering
	\resizebox{1.\textwidth}{!}
	{\Huge
	\begin{tabular}{p{8cm}p{10cm}p{10cm}p{8cm}p{30cm}}
		\hline
		\multicolumn{1}{c}{\textbf{Term}}             & \multicolumn{1}{c}{\textbf{Description}}                                                                                                                                                                                                                         & \multicolumn{1}{c}{\textbf{Methods}}    & \multicolumn{1}{c}{\textbf{Links to physics}}                                                                                                                              & \multicolumn{1}{c}{\textbf{Benefits and drawbacks}}                                                                                                                                                                                                                                                                                                                                                                                                       \\ \hline
		Physical Data-guided Machine Learning (PDgML) & PDgML is a supervised DL model that statistically learns the   known or unknown physics of a desired phenomenon by extracting features or   attributes from raw training data.                                                                                   & Classical physical data-guided neural networks (spatio-temporal   modeling, dynamic system modeling), deep operator networks,   physics-discovery neural networks. 
  & Learning from physical data (time series data of physical processes,  dynamic process data, and  physical process observations).   & \textbf{{[}Benefits{]}} Only   need simulations from physics model, fast inference, the discovery of  the dynamic behavior;  \newline \textbf{{[}Drawbacks{]}} The   architecture and learning process are black box, limited generalization and transferability.                                                                                                                                                                                                                       \\ \hline
		Physics-informed Machine Learning (PiML)      & PiML is a widely used approach to incorporating physical   constraints that can be trained from additional information obtained by   enforcing the physical laws (for example, designing loss functions   or regularization).                                    & Physics-informed neural networks (PINNs), physics-informed deep   neural operators (PINOs).     &   Integrating physical laws (e.g., parametrized PDEs) by  designing loss functions or regularization (e.g.,  PDE residual loss).                                                                   & \textbf{{[}Benefits{]}} Incomplete models and   imperfect data, strong generalization in the limited data, understanding   DL by the optimization process, tackling high dimensionality;   uncertainty quantification, data-physics-driven parameters  discovery;  \newline  \textbf{{[}Drawbacks{]}}  Effectiveness and   adaptability, data generation and benchmarks, large-scale applications, the   architecture and the learning process are still black boxes.\\ \hline
		Physics-embedded Machine Learning (PeML)      & PeML is achieved by embedding physics in the model frameworks or   modules.                                                                                                                                                                                       & Physical   equation-embedded machine learning (PEeML), physical property-embedded   machine learning (PPeML), physical condition-embedded machine learning (PCeML).    &  Embedding physics, including differential equations, physical properties (e.g., conservation, symmetry, causality), and physical conditions (e.g., boundary and initial conditions in PDEs, boundaries of physical processes) into ML frameworks or modules.
  &\textbf{{[}Benefits{]} } Interpretability of architecture and   learning process,  generalization; \newline   \textbf{{[}Drawbacks{]}} Computational cost is relatively high, hard inductive biases from   physics may hurt the representation power of neural nets, and require an in-depth   understanding of the physical principles to design.                                                                                                                               \\ \hline
		Physics-aware Hybrid Learning (PaHL)          & PaHL directly combines pure physics-based models, such as   numerical methods, climate, land, hydrology, and Earth system models with ML   models. Depending on the hybrid approach, hybrid learning can be categorized as serial, parallel, or complex. & Neural–numerical   hybrid learning, neural–hydrology hybrid learning.          &   Hybrid learning   by  combining   physics-driven models and   data-driven models.                                                                             & \textbf{{[}Benefits{]}} Can leverage both the flexibility of data-driven models along with the   interpretability and generalizability of physics-driven models, speed, and   operational convenience; \newline \textbf{ {[}Drawbacks{]}}   Need an optimal balance between the physics-driven and data-driven models.                                                                                                                                              \\ \hline
\end{tabular}}
\label{tab4}
\end{table}
\subsection{Physical Data-guided Machine Learning}

PDgML is a supervised DL model that statistically learns the known or unknown physics of a desired phenomenon by extracting features or attributes from raw physical data. 
\blue{These physical data typically encompass three aspects. First, there are time series data obtained from physical process models, including catchment or global scale rainfall-runoff time series data from datasets such as catchment attributes and meteorology for large-sample studies (CAMELS)~\cite{addor2017camels} and a global community dataset for large-sample hydrology (Caravan)~\cite{kratzert2023caravan}, among others, as well as climate and weather reanalysis data and products like ERA5~\cite{hersbach2020era5}. Second, these physical data include dynamic process data derived from physical dynamic models, such as various PDE numerical solution data like PDEBench~\cite{takamoto2022pdebench} and hydrodynamic data from flood dynamics simulations~\cite{xu2023large}. Finally, these physical data comprise other relevant physical observations for hydrological processes and events, such as remote sensing data, hydrological observations, and measurement data for floods and landslides~\cite{hydropml_dataset}. } Furthermore, 
PDgML consists of one or a combination of deep neural networks (DNN)~\cite{lecun2015deep}, convolutional neural networks (CNNs)~\cite{alzubaidi2021review}, recurrent neural networks (RNNs)~\cite{lipton2015critical}, graph neural networks (GNN)~\cite{scarselli2008graph}, generative adversarial networks (GAN)~\cite{creswell2018generative}, Transfomer~\cite{han2022survey}, deep reinforcement learning (DRL)~\cite{garnier2021review}, deep operator networks~\cite{kobayashi2023operator}, and physics-discovery neural networks~\cite{scellier2022agnostic}. The objectives and limitations of different physical data-guided ML are briefly discussed in Supplementary Table 1 of Supplementary Material (SM). 

\subsubsection{Classical Physical Data-guided Neural Networks}

\textbf{Spatio-temporal modeling.}
\blue{DL has become a powerful tool for solving complex problems involving large physical data. For the time series data obtained from physical process models, such as rainfall-runoff data, a series of time series modeling networks are proposed to extract features through supervised training.
To address the vanishing gradient issue inherent in RNNs, long short-term memory (LSTM) networks are developed~\cite{hochreiter1997long}.  These LSTM networks utilize three distinct gates to retain crucial information over extended periods while discarding what is deemed irrelevant. For modeling storage effects in hydrological processes, the advantage of LSTM networks is essential due to their ability to learn long-term dependencies between network inputs and outputs~\cite{kratzert2018rainfall}.  Physics-guided LSTM~\cite{xie2021physics} is developed to handle extreme events and monotonic relationships in rainfall-runoff simulations using DL. It incorporates three physical mechanisms to regulate the LSTM network, ensuring it effectively manages high flow, low flow, and monotonic properties.
However, current methods are designed within restricted problem settings, such as predicting fewer than 48 points~\cite{liu2020dstp}.
Recent studies highlight the potential of Transformer models to improve prediction capabilities. In response, Informer~\cite{zhou2021informer} has been introduced to address long sequence time-series forecasting, demonstrating the Transformer-like model's effectiveness in capturing long-range dependencies between inputs and outputs in lengthy sequences.}

Additionally, an army of physical data-guided spatio-temporal  models~\cite{shi2017deep,veillette2020sevir,wang2022predrnn} is proposed by imposing temporal and spatial inductive biases based on RNNs and CNNs. \blue{These physical data-guided spatio-temporal models integrate spatio-temporal data from physical models or observations, such as climate and weather reanalysis data and hydrological observations, with data-driven spatio-temporal models. These models achieve robust and accurate spatio-temporal predictions of physical variables, through effective supervised training.}
For example, ConvLSTM~\cite{shi2015convolutional} is a recurrent neural network designed for spatio-temporal prediction, incorporating convolutional layers in both the input-to-state and state-to-state transitions. DisasterNets~\cite{xu2023disasternets} is a type of CNN for spatio-temporal disaster mapping based on \blue{Earth observation data}. Furthermore, inspired by the strong representation capacity of Transformer architectures (such as bidirectional encoder representations from transformers (BERT)~\cite{vaswani2017attention} in natural language processing, vision Transformer in computer vision~\cite{dosovitskiy2020image, khan2022transformers}), quite a few physics-guided Transformer-based frameworks are presented based on \blue{physical simulation data}. For example, a space-time Transformer for Earth system forecasting, Earthformer~\cite{gao2022earthformer}, is presented based on an efficient space-time attention block. A global data-driven weather forecasting model, FourCastNet~\cite{pathak2022fourcastnet}, is proposed to provide accurate short- to medium-range global predictions at 0.25$\degree$ resolution. In addition, in order to improve prediction accuracy and uncertainty, irregular space-time grid prediction problems, and other issues, a collection of classic spatio-temporal models are proposed, such as GraphCast~\cite{lam2022graphcast} based on GNN for medium-range global weather forecasting, physically constrained GANs for improving local distributions and spatial structure of precipitation fields~\cite{hess2022physically}, and spatially-irregular forecasting based on geometric deep learning~\cite{rozemberczki2021pytorch}.

\textbf{Dynamic system modeling.}
Simulating multi-dimensional PDE systems with data-driven NNs has been a renewed research topic, that dates back to the last century~\cite{lee1990neural}. \blue{PDgML for dynamic system modeling  aims to learn a PDE solution (if available) from a provided simulation dataset and make accurate and efficient predictions using the learned model.}  Development has taken place in roughly three phases.

First, any early attempt~\cite{bongard2007automated}  at the data-driven discovery of hidden physical laws  compares numerical differentiations of the experimental data with analytic gradients of candidate functions. This approach employs symbolic regression and an evolutionary algorithm to determine the nonlinear dynamic system. Recently, motivated by the latest development of NNs, 
 a fully non-mechanistic method, PDE-Net~\cite{long2018pde}, is designed to predict the dynamics of complex systems accurately and uncover the underlying hidden PDE models based on learning convolution kernels (filters). 
\blue{Specifically, PDE-Net~\cite{long2018pde} utilizes the forward Euler method for temporal discretization, where each time step iteration is approximated to a nonlinear function using convolution operators with appropriately constrained filters. The nonlinear PDE system is then iteratively solved by stacking multiple convolution operators with shared weights.
 Further works have extended the approach by employing different networks to approximate the nonlinear function, including  residual CNNs~\cite{ruthotto2020deep}, symbolic neural networks~\cite{long2019pde},  feed-forward networks~\cite{xu2019dl}, and deep fully convolutional networks~\cite{lin2022universal}.}  
However, these models are fundamentally constrained by their reliance on discretizing the input domain using a sample-inefficient grid. As a result, they struggle to effectively manage temporally or spatially sparse and non-uniform observations, which are frequently encountered in practical applications. Recently, some convolution operators, such as  Mesh-Conv~\cite{hu2022mesh},  have been proposed to model  non-uniform structured or unstructured spatial mesh. \blue{Mesh-Conv~\cite{hu2022mesh} is achieved by introducing local weights to decompose standard convolutions, thereby integrating data structure information of spatial mesh.}

Second,  neural message passing-based models such as GNN-based model~\cite{sanchez2018graph}, contrastive learning-based message passing graph neural networks (MP-GNNs)~\cite{wangmessage}, and autoregressive solver~\cite{brandstetter2022message}, are related to interaction networks where the state of an object evolves as a function of its neighboring objects, forming dynamic relational graphs instead of grids.  Neural message passing is  a general framework for supervised (physical data-guided) learning on graphs~\cite{gilmer2017neural,brandstetter2022message}.  \blue{We illustrate the PDE solution process using the message passing neural PDE solver~\cite{brandstetter2022message} as an example. The discrete physical grid is constructed as a graph with nodes and edges, where nodes represent grid cells and edges define local neighborhoods. Employing the encode-process-decode framework for autoregressive solver, the encoder computes node embeddings, and the processor executes multiple time steps of learned message passing. Subsequently, the decoder generates predictions for the next timestep across spatial locations on the physical grid.}
In addition to the common mesh-based
 neural message passing-based approaches, particle-based frameworks have attracted considerable interest in scientific modeling for physical systems, particularly those employing GNNs~\cite{li2018learning, ummenhofer2020lagrangian,sanchez2020learning}.
Although neural message passing-based models can achieve geometry-adaptive learning of nonlinear PDEs with arbitrary domains, these models can apply message-passing between small-scale moving and interacting objects. 
 
 Third, DRL has been increasingly utilized in various data-driven dynamic system modeling due to its capability to address complex decision-making problems characterized by non-linearity and high dimensionality~\cite{novati2021automating,zheng2021data}. For instance, a DRL-based method~\cite{ma2018fluid} has been developed to manage a coupled 2D system with both fluid and rigid bodies, utilizing a position-based reward function. \blue{Initially, the fluid's velocity field is extracted using a convolutional autoencoder. Encoded velocity and other fixed features are subsequently fed into a multilayer perceptron (MLP) to determine actions. Both the MLP and autoencoder are trained using the DRL-based method~\cite{ma2018fluid}  to achieve physically realistic animations.}
 \subsubsection{Deep Operator Networks}
\blue{Deep operator networks constitute an emerging class of PDgML algorithms. Unlike traditional NNs that approximate solutions directly, these networks approximate mathematical operators, enabling faster solvers for dynamic processes represented by PDEs. Consequently, deep operator networks offer a robust alternative for extracting feature representations from dynamic process data, effectively addressing physical problems such as fluid dynamics equations and general PDE solutions.}
Operator learning strategies are based on the principles of the universal approximation theorem~\cite{lu2021learning}. This capability to approximate functions forms the foundation of modern developments in deep operator networks. Here, we present an introduction to deep operator networks (DeepONet)~\cite{lu2021learning}, the Fourier neural operator (FNOs)~\cite{li2020fourier}, the wavelet neural operators (WNOs)~\cite{tripura2022wavelet}, and the graph neural operators (GNOs)~\cite{li2020neural}.

DeepONets employ the aforementioned branch-trunk architecture to map finite inputs to the response space~\cite{lu2021learning}. 
This allows the trunk network to encode the domain, with multiple branch networks handling various inputs relevant to the problem. For instance, in~\cite{kontolati2023influence}, the decaying dynamics of the Brusselator reaction-diffusion system are incorporated by utilizing a trigonometric feature expansion of the temporal input in the trunk network. Additionally, to enable the application of DeepONet in realistic scenarios involving multiple input functions and diverse applications, a multiple input DeepONet~\cite{jin2022mionet} is introduced to handle multiple initial conditions and boundary conditions simultaneously.
 In addition, to enable the learning of the DeepONet with a smaller set of parameters, HyperDeepONet~\cite{lee2022hyperdeeponet} is proposed by using the expressive power of the hyper network.
However, several studies point out that the slow decay rate of the lower bound leads to inaccurate approximation operator learning for complex physical dynamics using DeepONet~\cite{kovachki2021neural,lanthaler2022error}. 

FNOs~\cite{li2020fourier}  use Fourier Transformations to move to the infinite-dimensional response space. These operators process input functions defined on a well-defined, equally spaced lattice grid and produce the desired fields at the same grid points. The network parameters are established and trained in the Fourier domain instead of the physical space. 
FNO has become one of the mainstream physical data-guided deep operator networks used to solve PDEs~\cite{kobayashi2023operator}.
Specifically, to solve PDEs on arbitrary geometries, Geo-FNO~\cite{li2022fourier} is designed by learning to deform the input (physical) domain. Domain agnostic FNO~\cite{liu2023domain} is proposed by incorporating a smoothed characteristic function within the integral layer architecture of FNOs. However, the FNO and the geo-FNO perform worse and are unstable on complex geometries and noisy data.  Thus, a host of variants of FNOs (such as factorized FNO~\cite{tran2021factorized}) are proposed to improve the generalization and stability of FNOs for solving PDEs.

A significant limitation of FNOs is that FFT basis functions are typically frequency localized without spatial resolution~\cite{bachman2000fourier}. Thus, WNOs are proposed, to learn network parameters in the wavelet space, which are both frequency and spatially localized, thereby enabling more effective learning of signal patterns. Recently, a coupled multiwavelets neural operator learning scheme~\cite{xiao2023coupled} is proposed to solve coupled PDEs, by decoupling the coupled integral kernels in the wavelet space.
This approach has demonstrated that WNO can effectively manage domains with both smooth and complex geometries. It has been utilized to learn solution operators for a highly nonlinear family of PDEs characterized by discontinuities and abrupt changes in both the solution domain and its boundary.

GNOs~\cite{li2020neural} utilize message passing on graph networks to capture the non-local structure of data. Graph kernel networks (GKNs), which originate from parameterizing Green’s functions in an iterative architecture~\cite{li2020neural}, are primarily employed in GNOs. Similar to FNOs, GKNs consist of a lifting layer, iterative kernel integration layers, and a projection layer. However, recent studies have noted that GKNs may exhibit instability with an increased number of iterative kernel integration layers. Therefore, a resolution-independent nonlocal neural operator~\cite{you2022nonlocal} has been proposed to address these issues.
\subsubsection{Physics-discovery Neural Networks}
Discovering dynamical processes is crucial as it enables us to uncover the fundamental physical laws that govern complex systems. However, the discovery of the governing equations presents a significant challenge due to the inherent complexity and nonlinearity of these systems, as well as the presence of noisy and incomplete data.
With the rapid advancement of  machine intelligence~\cite{wang2021physics}, the past three decades have witnessed significant development in physics-discovery neural networks, at the intersection of ML and scientific discovery. \blue{Physics-discovery neural networks facilitate the connection between data-driven models and physics learning by discovering and learning underlying physical principles and dynamic behaviors directly from complex physical data.}

Two main approaches have emerged based on unavailable and available prior knowledge of the physics system. First, the natural and optimal solution to automatically identifying the governing equations for dynamic systems is to learn a symbolic model from experimental data. Symbolic regression~\cite{schmidt2009distilling} and symbolic neural networks~\cite{sahoo2018learning,kim2020integration} have been explored extensively for inferring concise equations without prior knowledge. A physics-inspired method called AI Feynman~\cite{udrescu2020ai,udrescu2020ai2} is proposed for symbolic regression. 
AI Feynman employs NNs to discover generalized symmetries in complex data, enabling the recursive decomposition of difficult problems into simpler ones with fewer variables.
GNN~\cite{cranmer2020discovering} is also utilized to identify the nontrivial relation due to its exceptional inductive capability. In addition, PDgML is applied for knowledge discovery  that cannot be
modeled from a physical process-based perspective, such as the two-way feedback between human and water systems~\cite{wang2021predicting, meempatta2019reviewing}.
Second, discovering dynamic behavior from data often involves defining a large set of possible mathematical basis functions or process-based models based on prior knowledge. Representative works include  Sparse Identification of Nonlinear Dynamics (SINDy)~\cite{brunton2016discovering} and PDE functional identification of nonlinear dynamics (PDE-FIND)~\cite{rudy2017data} for ordinary differential equations (ODEs) and PDEs, respectively. However, these standard sparse representation-based methods are typically limited to high-fidelity noiseless measurements, which are often difficult and expensive to obtain. Consequently, recent efforts have focused on discovering PDEs from sparse or noisy data~\cite{xu2021deep}. 


\subsection{Physics-informed Machine Learning}
In scientific computing, physical phenomena are typically described using a robust mathematical framework that includes governing differential equations, as well as initial and boundary conditions.
PiML is a widely-used approach that integrates physical laws into ML models (for example, designing loss functions or regularization), facilitates the accurate capture of dynamic patterns and concomitantly diminishes the search space for model parameters.
This approach is sometimes referred to as imposing differentiable constraints in loss functions.   It integrates (noisy) data and mathematical models, and implements them through neural networks (physics-informed neural networks) or kernel-based neural operators (physics-informed neural operators). The objectives and limitations of different physics-informed machine learning are briefly discussed in Supplementary Table 2 of SM.

\subsubsection{Physics-informed Neural Networks}

PDE solutions are represented as NNs by including the square of the PDE residual in the loss function, resulting in a NN-based PDE solver~\cite{lagaris1998artificial}. Recently, this approach is refined further and called ``physics-informed neural networks (PINNs)~\cite{raissi2019physics},” initiating a flurry of follow-up work. For example, we consider a parametrized PDE system given by
\begin{equation}
\begin{gathered}
f\left(\mathbf{x}, t, u, \frac{\partial u}{\partial \mathbf{x}}, \frac{\partial^{2} u}{\partial \mathbf{x^{2}}}, \frac{\partial u}{\partial t}, \ldots, \vartheta\right)=0, \quad \mathbf{x} \in \Omega, t \in[0, T], \\
u\left(\mathbf{x}, t_{0}\right)=g_{0}(\mathbf{x}) \quad \mathbf{x} \in \Omega, \quad
u(\mathbf{x}, t)=g_{\Gamma}(t) \quad \mathbf{x} \in \partial \Omega, t \in[0, T],
\end{gathered}
	\label{eq1}
\end{equation}
where $u: [0, T] \times \mathbf{X} \rightarrow \mathbb{R}^{n}$ is the solution, subject to initial condition $g_{0}(\mathbf{x})$ and boundary condition $g_{\Gamma}(t)$, which can be of various types such as periodic, Dirichlet, or Neumann. The PDE parameters are denoted by $\vartheta$ = [$\vartheta1$, $\vartheta2$, . . . ]. The residual of the PDE is represented by $f$, which includes the differential operators (i.e., $ \frac{\partial u}{\partial \mathbf{x}}, \frac{\partial^{2} u}{\partial \mathbf{x^{2}}}, \frac{\partial u}{\partial t}$, . . . ). The physical domain is denoted by $\Omega$, with its corresponding boundary represented by $\partial \Omega$.

Vanilla PINNs~\cite{raissi2019physics} directly approximate the solution of differential equations  in a physics-informed fashion with less training data~\cite{sun2020physics} or without any training data~\cite{sun2020surrogate,  jin2021nsfnets, bihlo2022physics}. 
In PINNs, solving a PDE system is converted into an optimization problem by minimizing the loss function to iteratively update the NN,
\begin{equation}
\begin{gathered}
\min _{\theta} \mathcal{L}_{\mathrm{PINN}}(\theta)=w_i \mathcal{L}_{I C}+w_b \mathcal{L}_{B C}+w_d \mathcal{L}_{\text {Data }}+w_p \mathcal{L}_{P D E}, \\
\mathcal{L}_{I C} = \|u\left(\mathbf{x}, t_{0}\right) - g_{0}(\mathbf{x}) \|_{\Omega}, \quad \mathcal{L}_{B C} = \|u(\mathbf{x}, t) - g_{\Gamma}(t)\|_{\partial \Omega}, \quad
\mathcal{L}_{\text {Data }} = u|_{\Omega}-\hat{u}|_{D a t a}, \\
\mathcal{L}_{P D E} = \|f\left(\mathbf{x}, t, u, \frac{\partial u}{\partial \mathbf{x}}, \frac{\partial^{2} u}{\partial \mathbf{x^{2}}}, \frac{\partial u}{\partial t}, \ldots, \vartheta\right)\|_{\Omega},
\end{gathered}
\label{eq2}
\end{equation}
where $w_i, w_b, w_d, w_p$ are  regularization parameters that control the emphasis on residuals of initial conditions ($\mathcal{L}_{I C}$), boundary conditions ($\mathcal{L}_{B C}$), training data ($\mathcal{L}_{\text {Data }}$), and  PDE ($\mathcal{L}_{P D E}$), respectively. $\mathcal{L}_{\text {data }}$ measures the mismatch between the NN prediction $u$ and the training data $\hat{u}$. The NN parameters are denoted by $\theta$, which takes the spatial and temporal coordinates (x, t), and possibly other quantities, as inputs and then outputs $u$.
\begin{figure}[!htp]
	\centering
	{\includegraphics[width = 0.9\textwidth]{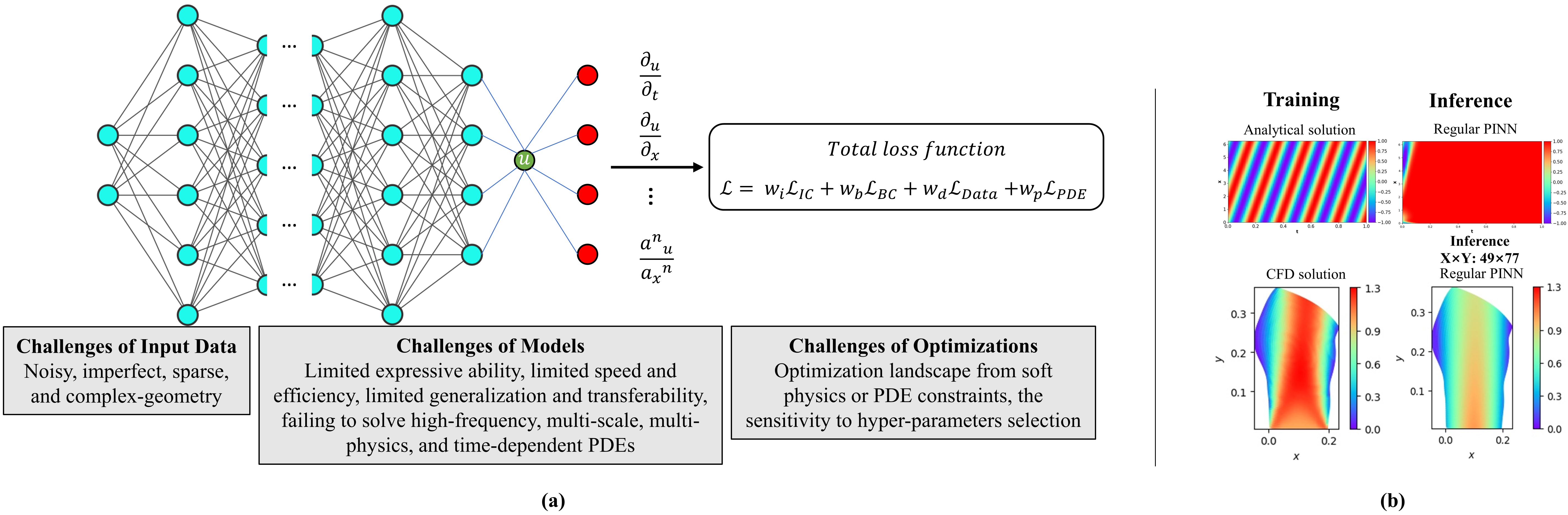}}
	\caption{(a) Challenges of vanilla PINNs. (b) An example of vanilla PINNs failing to converge on high-frequency and multi-scale PDEs. 1-D convection equations with high-frequency (first row): Train and infer on a  spatial-temporary resolution $256 \times 100$; 2-D steady incompressible Navier-Stokes equations (second row): Train and infer on a  spatial resolution $49 \times 77$. Analytical solution and computational fluid dynamics (CFD) represent ground truth.}
	\label{fig:1}
\end{figure}

In current ML-based methods for scientific computing, vanilla PINNs are typically formulated in a pointwise manner, as depicted in Fig.~\ref{fig:1}(a). These methods utilize NNs as the ansatz for the solution function and optimize a loss function to minimize violation of the given equation, leveraging auto-differentiation for exact, mesh-free derivatives. One of the primary advantages of vanilla PINNs is their ability to avoid the need for a discretization grid, making them particularly useful for solving inverse and higher-dimensional problems.
However, despite these advantages, vanilla PINNs exhibit lower efficiency compared to classical methods in solving most PDEs, for several reasons~\cite{fuks2020limitations, wang2022and}:
(1) the challenging optimizations due to soft physics or PDE constraints; (2) the difficulty of conveying information from the initial or boundary conditions to unseen interior regions or to future times; (3) the sensitivity to hyperparameter selection; (4) the slow training/re-optimization process in practical applications and different instances; and (5) failure to converge on time-dependent PDEs or high-frequency, multi-scale PDEs, as shown in Fig.~\ref{fig:1}(b).
 
As elucidated in Supplementary Table 2 of SM, a vast and growing body of work aims to address these challenges. Specifically, it encompasses the following eight aspects. First, many problem-specific insights are utilized by improving the training of PINNs. For example, Universal PINNs~\cite{huang2022universal} improve the PINNs by a lower bound constrained uncertainty weighting algorithm and a multi-scale deep NN. A distinct  conservative physics-informed neural network (cPINN)~\cite{jagtap2020conservative} is developed for discrete domains dealing with nonlinear conservation laws. Another approach, nPINNs~\cite{pang2020npinns}, extends PINNs to handle parameter and function inference for integral equations like nonlocal Poisson and nonlocal turbulence models. An improved PINN approach~\cite{pu2021solving} is presented to solve localized wave solutions of the derivative nonlinear Schrödinger equation in complex space, achieving fast convergence and optimal simulation performance. An automatic numerical solver~\cite{wight2020solving} is proposed for the Allen-Cahn and Cahn-Hilliard equations employing advanced PINNs. A simple  and efficient
 characteristic-informed neural network~\cite{braga2022characteristics} is developed for solving forward and inverse problems in hyperbolic PDEs. NSFnets~\cite{jin2021nsfnets} is proposed for the incompressible Navier-Stokes equations using automatic differentiation. 
 Furthermore, 
 for solving Navier-Stokes equations on irregular geometries,
  a geometry-aware physics-informed neural network (GAPINN)~\cite{oldenburg2022geometry} is presented. 
 In addition, a coupled physics-informed neural networks (CPINNs)~\cite{zhang2023cpinns} is designed for closed-loop geothermal systems.

Second, PINNs are not equipped with
input data preprocessing, which may restrict their application, especially
for scenarios where the input data are noisy, imperfect, or sparse, or geometry  is complex. Bearing these concerns in mind, 
Bayesian-PINNs~\cite{yang2021b} integrate physical laws with scattered noisy measurements to deliver predictions while quantifying aleatoric uncertainty within a Bayesian framework. An extension of PINNs with enforced truncated Fourier
decomposition (ModalPINNs)~\cite{raynaud2022modalpinn} is proposed for  flow reconstruction using imperfect and sparse information. In addition, a positional encoding mechanism for PINNs based on the eigenfunctions of the Laplace-Beltrami operator ($\Delta$-PINNs)~\cite{costabal2022delta} is presented for complex geometries. 

Third, due to the limited expressive capacity inherent in fully-connected networks within vanilla PINNs,  the training process of vanilla PINNs may lack robustness and stability, potentially hindering convergence to the global minimum. 
 In addition, fully-connected networks struggle to learn multiscale and multiphysics problems. \blue{To resolve this issue, more robust NN architectures, such as a multi-scale deep neural network in universal PINNs~\cite{huang2022universal},  an ensemble model
 of PINNs~\cite{haitsiukevich2022improved}, graph neural networks in physics-informed GNNs~\cite{yan2022physics}, multi-head physics-informed neural networks (MH-PINNs)~\cite{zou2023hydra}, a convolutional backbone in Phase2vec~\cite{ricci2022phase2vec}, and orthogonal polynomials in PI-PINN~\cite{tang2023physics}  have been developed. In addition, researchers start to blend physics with RNN-based, CNN-based, or GNN-based learning. For example,
 a physics-informed convolutional-recurrent learning architecture (PhyCRNet)~\cite{ren2022phycrnet} is proposed for solving PDEs without any labeled data. A physics-constrained CNN learning architecture, PhyGeoNet~\cite{gao2021phygeonet}, is developed to learn solutions of parametric PDEs on irregular domains without relying on labeled data.
  A fast and continuous CNN-based  solution (Spline-PINN)~\cite{wandel2022spline} is designed for approaching PDE training without any precomputed training data. Graph network~\cite{michelis2022physics} is used to represent meshes, and physics-based loss is formulated to provide an unsupervised learning framework for PDEs. These models have advanced in exploring network frameworks and effective feature representations. However, they usually have limited accuracy due to gradient solving at the grid scale (such as finite difference). }
 
Fourth, vanilla PINNs tend to be significantly slower than classical numerical methods. To address this, meta-learning based methods have been proposed for efficiently solving a variety of related PDEs using PINNs. For instance, de Avila Belbute-Peres et al.~\cite{de2021hyperpinn} meta-train a hypernetwork that can generate weights for a small NN specific to each task, providing an approximate solution to the PDE. Psaros et al.~\cite{psaros2022meta} meta-learn a loss function that optimizes the NN, achieving performance benefits over hand-crafted and online adaptive loss functions. Penwarden et al.~\cite{penwarden2021physics} propose a meta-learning approach similar to Meta-PDE approach, which learns an initialization of weights to enable quick optimization of the NN. Penwarden et al.~\cite{penwarden2021physics} find that Model-Agnostic Meta-Learning (MAML)~\cite{finn2017model} achieves poor performance, only marginally better than random initialization; the recent work~\cite{qin2022meta} comes to the opposite conclusion, that MAML-based Meta-PDE performs well for a given runtime. In addition, a cohort of other methods is also proposed to improve the efficiency of PINNs. For instance, XPINNs~\cite{jagtap2021extended} is presented by  space and time parallelization. To replace the computationally expensive automatic differentiation in PINNs, meshless radial basis function-finite differences in DT-PINNs~\cite{sharma2022accelerated} are applied by sparse-matrix vector multiplication, and a coupled-automatic-numerical differentiation framework (canPINN)~\cite{chiu2022can} unifies the advantages of automatic differentiation and numerical differentiation, providing more robust and efficient training than automatic differentiation-based PINNs.

Fifth, to improve the generalization of PINNs, a transfer PINN~\cite{manikkan2022transfer} is used to solve forward and inverse problems in nonlinear PDEs by parameter sharing. Sixth,  PINNs encounter challenges when approximating target functions with high-frequency or multi-scale features. A sea of variants of PINNs are proposed to address this problem. For example, finite basis PINNs~\cite{dolean2022finite} use ideas from domain decomposition for complex, multi-scale solutions. Krishnapriyan et al.~\cite{krishnapriyan2021characterizing} use curriculum PINN regularization and sequence-to-sequence learning to address the problems of PINNs that fail to capture relevant physical phenomena for slightly more complex problems. In addition, spatio-temporal and multiscale random Fourier features in multiscale PINNs~\cite{wang2021eigenvector} are used to address the high-frequency or multi-scale problems in PINNs. Seventh, adaptive algorithms for weighing components of the PINN loss function (hyperparameter selection) have also been proposed, such as self-adaptation PINNs~\cite{mcclenny2020self}, a heuristic method~\cite{van2022optimally}.

Finally, PINNs are employed to address the inverse problem of parameter identification, also known as data-physics-driven parameter discovery. The inverse problem of parameter estimation is often ill-posed due to the non-uniqueness resulting from a large number of unknowns. Moreover, the stability of solutions against data noise and modeling errors is typically not guaranteed. Physics-Informed Neural Networks with Functional Connections (PINN-TFC), such as the extreme theory of functional connections~\cite{schiassi2021physics}, have been developed to mitigate these challenges. PINNs with sparse regression~\cite{chen2021physics} have also been proposed for discovering the governing PDEs of nonlinear spatiotemporal systems from sparse and noisy data. This discovery method integrates the strengths of NNs in rich representation learning, physical embedding, automatic differentiation, and sparse regression. Saqlain et al.~\cite{saqlain2022discovering} select a series of increasingly complex but physically relevant examples and explore the use of PINNs in intrinsically discrete, high-dimensional settings. However, solving such inverse problems using PINNs is still at an early stage, and further studies are certainly warranted.
\subsubsection{Physics-informed Neural Operators}
To overcome the challenges of vanilla PINNs, physics-informed neural operators (PINOs)~\cite{li2021physics} are proposed. These utilize both the data and equation constraints (whichever are available) for deep operator networks~\cite{rosofsky2023applications}. 

One particular PINO, physics-informed DeepONet~\cite{wang2021learning},  trains the DeepONet by integrating known differential equations directly into the loss function along with labeled dataset information. 
The output of DeepONet is differentiable with respect to its input coordinates, enabling the use of automatic differentiation to develop effective regularization mechanisms that bias the target output function towards satisfying the underlying PDE constraints. However, the PDE loss of physics-informed DeepONet is computed at any query point. The input sensors limit this PINO to a fixed grid or basis, and therefore it is not discretization invariant. In addition, its architecture comes with the limitations of linear approximation. Furthermore, in order to enhance the computational efficiency and generalization of physics-informed DeepONet, a subset of problem-specific  surrogate models based on the physics-informed DeepONet  is designed, such as physics-informed DeepONet for chemical kinetics~\cite{zanardi2022adaptive}, and physics-informed variational formulation
of DeepONet for brittle fracture analysis~\cite{goswami2022physics}. Furthermore, in order to achieve discretization invariance, the physics-informed FNO (PINO)~\cite{li2021physics} is proposed by  integrating operator learning and physics-informed settings. PINO reduces the need for labeled datasets during neural operator training and facilitates faster convergence of solutions.  In this setting, it's crucial to highlight that, unlike DeepONet, FNO produces the solution on a grid using spectral methods, finite difference, or other numerical gradient solution methods. However, PINO has limited accuracy in solving non-periodic problems due to the numerical gradient method and relies on certain training samples to achieve better accuracy. In addition, the adaptability to different PDE solution problems is not strong due to the selection of model parameters. It cannot be applied to large-scale and long-term sequence PDE solutions. In order to address the problem that PINO cannot represent nonperiodic functions, an architecture that leverages Fourier continuation (FC-PINO)~\cite{maust2022fourier} is proposed to apply the exact gradient method to PINO. In addition, a geometry-adaptive physics-informed neural solver (GeoPINS)~\cite{xu2023large} is proposed based on the advantages of no training data in physics-informed neural networks; this solver also possesses a fast, accurate, geometry-adaptive and resolution-invariant architecture. Most recently, 
a physics-informed WNO~\cite{tripura2023physics} is proposed for learning the solution operators of families of parametric PDEs without the need for labeled training data, leveraging the advantage of time-frequency localization of wavelets.

\subsection{Physics-embedded Machine Learning}
PeML is achieved by embedding physics in the model frameworks or modules. It can be divided into physical equation-embedded ML (PEeML), physical property-embedded ML (PPeML), and physical condition-embedded ML (PCeML).

\subsubsection{Physical Equation-embedded Machine Learning}
PEeML uses the knowledge of specific equations (such as differential equations) to design the frameworks or modules of ML. The objectives and limitations of different PEeML are briefly discussed in Supplementary Table 3 of SM.

Specifically, a powerful framework, AutoIP~\cite{long2022autoip}, is introduced to automatically integrate physics into Gaussian processes, thereby improving prediction accuracy and uncertainty quantification across diverse differential equations. Additionally, a physics-aware finite volume neural network (FINN)~\cite{karlbauer2022composing} is designed for modeling spatio-temporal advection-diffusion processes by compositively representing the elements of PDEs.
Deep Lagrangian networks (DeLaN)~\cite{lutter2019deep} encode the Euler-Lagrange equation derived from Lagrangian mechanics. DeLaN can be optimized end-to-end while ensuring adherence to physical principles.
A framework called MeshGraphNets~\cite{pfaff2020learning} is proposed to solve the underlying PDEs using graph neural networks. MeshGraphNets works by encoding the simulation state into a graph, performing calculations in both the mesh space (defined by the simulation mesh) and the Euclidean world space (where the simulation manifold is embedded). This approach allows the approximation of differential operators essential for the internal dynamics of most physical systems. 
Furthermore, 
fluid graph networks (FGN)~\cite{li2022graph}
 using graphs to represent the fluid field, is proposed for solving a Lagrangian representation of the Navier-Stokes equation.
FGN maintains key physical properties of incompressible fluids, such as low-velocity divergence. Several follow-up works are introduced using graph neural networks~\cite{fortunato2022multiscale,halimi2023physgraph}.
Apart from building an explicit graph structure, an efficient ConvNet architecture based on a continuous convolution layer~\cite{ummenhofer2020lagrangian} is developed. The network processes sets of particles in which dynamic particles and static particles are used to represent the fluid and describe the boundary of the scene, respectively. 

Furthermore, the use of differentiable modules in NNs is a promising direction for physical equation-embedded ML. It implements physical equations as differentiable feature space or computational graphs~\cite{hu2019difftaichi}, enabling the optimization of dynamic processes with analytical gradients and therefore improving sample efficiency. For example, a differentiable physics-informed graph network (DPGN)~\cite{seo2019differentiable} is proposed to incorporate implicit physics equations in latent space. A differentiable layer~\cite{negiar2022learning}, called PDE-Constrained-Layer, is developed by using implicit differentiation, thereby allowing us to train NNs with gradient-based optimization methods. A spatial difference layer~\cite{seo2020physics} is utilized in physics-aware difference graph networks (PA-DGN) to efficiently exploit neighboring information under the limitation of sparsely observable points. A physics-based finite difference convolution connection~\cite{rao2023encoding} in a  physics-encoded recurrent CNN is introduced to facilitate the learning of the spatiotemporal dynamics in sparse data regimes.
Furthermore, the term differentiable modeling is proposed to include any method that can produce gradients rapidly and accurately at scale, potentially serving as the basis for unifying NN and process-based geoscientific modeling~\cite{shen2023differentiable}. 

\subsubsection{Physical Property-embedded Machine Learning}
Physical properties usually include conservation, symmetry, and causality of physical systems.  
By building a NN that inherently respects a given physical property, we thus make conservation of the associated quantity more likely and consequently the model’s prediction more physically accurate. 

\textbf{Conservation.}
Many real-world systems adhere to conservation laws related to mass, energy, momentum, or particle number, typically expressed through continuity equations. For example, in hydrology, it is the amount of water~\cite{beven2011rainfall}. However, standard DL methods, such as CNNs and LSTMs, encounter challenges in maintaining conservation across layers or time steps. To address this issue, the mass-conserving LSTM (MC-LSTM)~\cite{hoedt2021mc} extends the inductive bias of the LSTM to uphold these conservation laws, ensuring that mass input is conserved through modifications to the recurrent structure of the traditional LSTM. The main concept is to utilize memory cells from LSTMs as mass accumulators or storage units. In addition, an antisymmetrical continuous convolutional layer~\cite{prantl2022guaranteed} in a hierarchical network is presented by enforcing the conservation of momentum with a hard constraint. 
The parameterization of deep neural networks~\cite{richter2022neural} is tailored to meet the continuity equation, ensuring the adherence to fundamental conservation laws by creating divergence-free NNs. An implicit neural network layer~\cite{smith2022physics} that incorporates the conservation of mass is proposed by using implicit differentiation. Furthermore, a structured approach to enforce nonlinear analytic constraints such as energy and mass conservation within NNs~\cite{beucler2021enforcing} is proposed. The concept of ``conversion layers'' is introduced in the architecture that transforms nonlinearly constrained mappings into linearly-constrained mappings within NNs without overly degrading performance.

\textbf{Symmetry} is implicitly leveraged in DL frameworks to design networks with invariant and equivariant properties. Let  $f: X \rightarrow Y$ be a function and  $G$ be a group. Assume $G$ acts on $X$ and $Y$. If $f(g x)=g f(x)$ for all $x \in X$ and $g \in G$, 
the function $f$ is $G$-equivariant. If $f(g x)=f(x)$ for all $x \in X$ and $g \in G$, the function $f$ is G-invariant.

CNNs enabled breakthroughs in computer vision by leveraging translation equivariance. Similarly, RNNs, GNN, and capsule networks all impose symmetries~\cite{wang2021physics}. There is a deep connection between symmetries and physics. For instance, Noether's theorem~\cite{kosmann2011noether} establishes a relationship between conserved quantities and symmetry groups. Moreover, symmetries in fluid dynamics~\cite{wang2020incorporating} are employed in the design of equivariant networks. The Navier-Stokes equations and the heat equation exhibit invariance under the following transformations: 

- Space translation: $T_c^{\mathrm{sp}} w(x, t)=w(x-c, t), \quad c \in \mathbb{R}^2$,

- Time translation: $T_\tau^{\text {time}} w(x, t)=w(x, t-\tau), \quad \tau \in \mathbb{R}$,

- Uniform motion: $T_c^{\mathrm{um}} w(x, t)=w(x, t)+c, \quad c \in \mathbb{R}^2$,

- Rotation/Reflection: $T_R^{\text {rot }} w(x, t)=R w\left(R^{-1} x, t\right), R \in O(2)$,

- Scaling: $T_\lambda^{s c} w(x, t)=\lambda w\left(\lambda x, \lambda^2 t\right), \quad \lambda \in \mathbb{R}_{>0}$.

Recently, incorporating symmetries of physics into NNs has been developed in the context of CNNs~\cite{lang2020wigner}, GNNs~\cite{han2022geometrically}, and neural operators~\cite{liu2023ino}.

First,
an end-to-end ML-based deep potential–smooth edition (DeepPot-SE)~\cite{zhang2018end} is developed, which is extensive, continuously differentiable, and scales linearly with system size while preserving all natural symmetries. The symmetry preserving functions maintain translational, rotational, and permutational symmetries. 

Second, there have been a number of GNN-based works, called geometrically equivariant graph neural networks~\cite{han2022geometrically}, which leverage symmetry as an inductive bias in learning simulations. These models ensure that their outputs rotate, translate, or reflect in the same manner as their inputs, preserving symmetry.For example, EGNN~\cite{satorras2021n} models interactions using invariant distances, calculated via inner products of relative positions, while GMN~\cite{huang2022equivariant} extends to a multi-channel version with a stack of vectors. Subequivariant GNN~\cite{han2022learning} relaxes the condition of full equivariance to subequivariance, by considering external fields like gravity and retaining the theoretical ability for universal approximation. 

Third, an integral neural operator architecture~\cite{liu2023ino} is designed to learn physical models with fundamental conservation laws automatically guaranteed. The 
translation- and rotation-invariant neural operator is developed by replacing the frame-dependent position information with its invariant counterpart in the kernel space; consequently it abides by the conservation laws of linear and angular momenta. A group equivariant  FNO~\cite{helwig2023group} is introduced, featuring Fourier layers designed to encode symmetries by exhibiting equivariance to rotations, translations, and reflections within the neural operator architecture. 

\textbf{Causality.} A fundamental pursuit in physics is to identify causal relationships. Integrating causality
into ML promises to advance the comprehension of physical processes and bolster the robustness of ML models~\cite{runge2023causal}.
For example, causality-DeepONet~\cite{liu2022causality} implements the physical causality in the DeepONet structure. Specifically, the causality-DeepONet is employed to learn operators that capture the response of buildings to earthquake ground acceleration. However, many challenges in causal embedding in ML remain unresolved. These include leveraging causality to enhance DL models, understanding system responses under interventions, disentangling complex and multiple processes, and designing environments to control physical phenomena. Additionally, a causal training algorithm~\cite{wang2022respecting} has been introduced to restore physical causality during PINNs model training by appropriately reweighting the PDE residual loss in each iteration of gradient descent. This straightforward adjustment enables PINNs to address complex problems such as the incompressible Navier-Stokes equations in turbulent regimes.

\subsubsection{Physical Condition-embedded Machine Learning}
\blue{There are many physical conditions in the physical system, such as boundary conditions and initial conditions in PDEs, boundaries of land use and land cover in the physical environment, and so on. Failing to satisfy these conditions can result in unstable models and non-physical solutions. Therefore, it is essential for ML models to adhere to these constraints. Physical condition-embedded ML establishes a connection with physical learning by embedding these physical conditions into ML frameworks or modules, ultimately accurately capturing underlying physical principles and providing reliable models for scientific and engineering applications.}
Specifically, for PDEs, three types of boundary conditions (BCs) that are widely used to model physical phenomena are (1) Dirichlet: fixed-value at the boundary; (2) Neumann: fixed-derivative at the boundary; and (3) periodic: equal values at the boundary. To integrate boundary conditions into NNs, a boundary enforcing operator network~\cite{saad2022guiding} is introduced. This network ensures the satisfaction of boundary conditions by modifying the operator kernel's structure. The physics-embedded neural network~\cite{horie2022physics} is developed by considering BCs in GNN.  
\subsection{Physics-aware Hybrid learning}
\begin{figure}[!htp]
	\centering
	{\includegraphics[width = 0.9\textwidth]{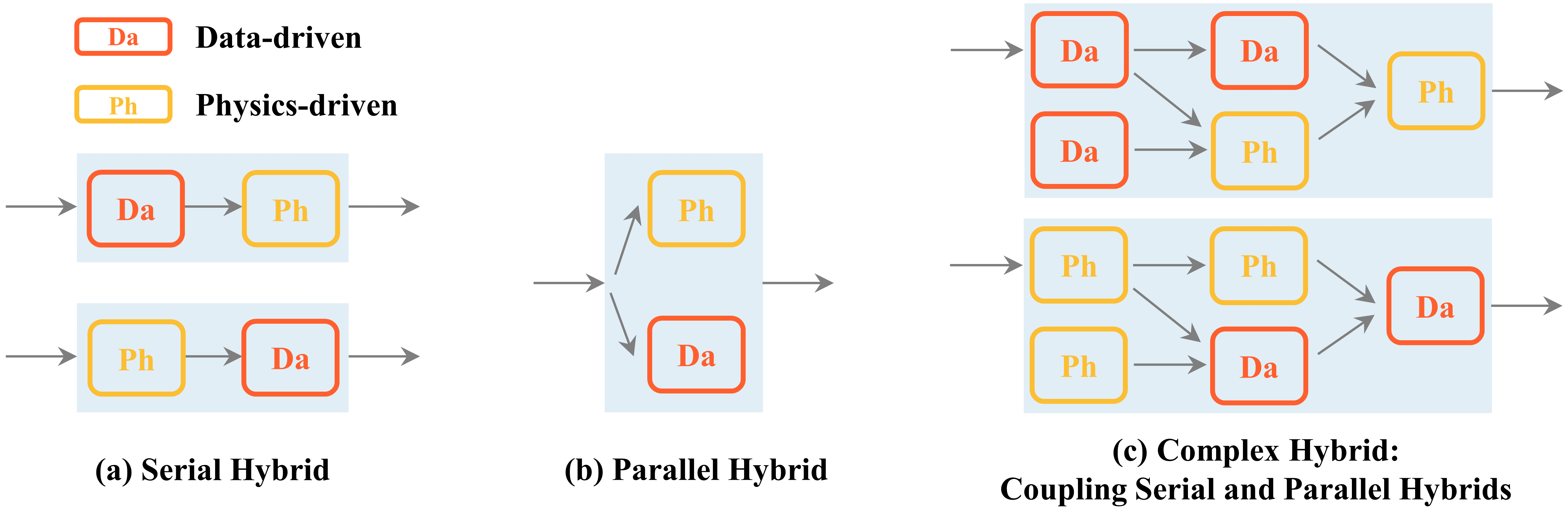}}
	\caption{\blue{Different hybrid approaches in physics-aware hybrid learning.}}
	\label{fig:0}
\end{figure}

PaHL directly combines pure physics-based models, such as numerical methods, climate, hydrology, land, and Earth system models, with ML models. As shown in Fig.~\ref{fig:0}, according to the hybrid way, hybrid learning can be divided into serial, parallel, and complex ways of approaching learning.
\subsubsection{Neural–numerical Hybrid Learning}
PDEs are notoriously difficult to solve, and traditional numerical approximation schemes have high computational costs. Recently, hybrid neural-numerical solvers have been developed to complement the modern trend of fully end-to-end learning systems.

\textbf{Serial Hybrid}  has broad potential advantages for physical models and data-driven models. For example,  NNs are used to handle data coming from irregular domains.
 A family of neural ODEs~\cite{chen2018neural,dupont2019augmented} transforms traditionally discretized neuron layer depths into continuous equivalents by parameterizing the derivative of the hidden state with a NN. The network's output is then computed using a black-box differential equation solver, effectively merging NNs with numerical solvers in a serial hybrid manner. In addition,  the advantages of  physics solvers (such as their scale-invariant properties) can be  imparted into NNs. 
 A novel hybrid training approach~\cite{holl2022scale} that combines  a scale-invariant physics solver with ML techniques is developed for solving inverse problems, like parameter estimation and optimal control. Gradients of NNs are replaced by the update computed from a higher-order solver that can encode the scale-invariant property of the physics. \blue{Furthermore, a serial hybrid unit combining a discrete physics solver and a NN~\cite{fan2024differentiable} is proposed to solve complex fluid-structure interaction problems. The physics solver encodes the control PDEs in discrete form as a non-trainable component, which is seamlessly integrated with a trainable LSTM network to form the entire recurrent unit. The hybrid model is trained by considering both fluid and solid physics dynamics as loss functions. Experimental results demonstrate that the serial hybrid model outperforms purely data-driven neural models.}

\textbf{Parallel Hybrid} methods are designed to develop fast, robust, and reliable solvers for solving PDEs. Physics-driven models are usually used as a supervised signal to train data-driven NNs.
For example, a solver named HINTS (Hybrid, Iterative, Numerical, and Transferable Solver)~\cite{zhang2022hybrid} is proposed for differential equations. HINTS combines traditional relaxation techniques (such as Jacobi, Gauss-Seidel, conjugate gradient, multigrid methods, and their variants) with DeepONet. This approach offers faster solutions for a wide range of differential equations while maintaining machine-level accuracy.  
A fast, differentiable hybrid approach~\cite{nava2022fast} integrates a 2D direct numerical simulation for deformable solid structures with a physics-constrained neural network surrogate to capture fluid hydrodynamic effects. \blue{A hybrid model enhanced by the finite element method~\cite{meethal2023finite} is proposed to create a high-performance surrogate model for forward and inverse PDE problems. This model hybridizes the NN and the finite element method in parallel to output the physical quantities used to calculate the loss function (residual vector). Additionally, parallel hybrid allows numerical solvers to achieve high accuracy on coarse grids, significantly reducing computational time. For instance, a deep neural network multigrid solver~\cite{margenberg2024dnn} enhances the finite element method on coarse grids by utilizing fine-scale information from NNs. This hybrid method is realized through the interaction between the finite element method and NNs on local grids.}

\blue{\textbf{Complex Hybrid (Coupling Serial and Parallel Hybrids)} methods are flexible frameworks that can be used to address many problems in solving PDEs. As shown in Fig.~\ref{fig:0}(c) top, data-driven models can be integrated into the input space by serial hybrid and the internal space by parallel hybrid.
For example, in order to address the scalability issue of solving PDEs on a large scale, a physics-aware downsampling method~\cite{giladi2021physics} is proposed  by minimizing the distance between the solutions on the fine and coarse grids. Specifically, a serial hybrid method is designed to downsample fine grid terrain, by combining a numerical solver (such as finite difference) on coarse grid terrain and a downsampling neural network. Then, a parallel numerical solver on fine grid terrain is used to provide a supervised signal for the predictions on the coarse grids. In addition, as shown in Fig.~\ref{fig:0}(c) below, data-driven models can be integrated into the internal space by parallel hybrid and the output space by serial hybrid. For instance, a hybrid graph neural network~\cite{belbute2020combining} that merges a traditional graph convolutional network (GCN) with an integrated differentiable fluid dynamics simulator has been developed. Through parallel hybrid, fine grid parameters are input to the GCN to obtain the graph representation, while the coarse grid parameters are input to the fluid dynamics simulator to generate simulation results. The upsampled simulation results from the fluid dynamics simulator are subsequently fed into a GCN through serial hybrid, ultimately predicting the desired output values. This complex hybrid model generalizes well to new scenarios and benefits from the significant speedup of neural network-based CFD predictions, significantly outperforming standalone coarse CFD simulations. }

\subsubsection{Neural–hydrology Hybrid Learning}
The hydrological model is a standard representation of physical processes, serving as an input-output model used to simulate the evolution and dynamics of surface and groundwater storage, fluxes, and physical properties of the Earth~\cite{devia2015review}.  
Neural-hydrology hybrid learning integrates various hydrological models using ML methods to produce a final process-based prediction product. This approach has garnered increased attention recently due to its ability to enhance the interpretability of hydrological models, improve understanding of ML techniques, and leverage advances in computational resources and methods~\cite{slater2023hybrid}. 
Here we list some recent neural–hydrology hybrid methods for hydrological variable prediction, such as streamflow and runoff.

\textbf{Serial Hybrid} integrates hydrological models and data-driven models in a sequential manner. For instance, the hybrid monthly runoff forecasting method~\cite{humphrey2016hybrid} utilizes simulated soil moisture from the GR4J conceptual rainfall-runoff model to represent initial catchment conditions within a Bayesian neural network framework. This serial hybrid model demonstrates superior performance compared to both the standalone GR4J model and the Bayesian neural network. Additionally, to mitigate the computational complexity associated with parallel hybridization of conceptual rainfall-runoff models and ML techniques, Okkan et al.~\cite{okkan2021embedding} embeds NNs and support vector regression into a monthly centralized conceptual rainfall-runoff model. Furthermore, Mohammadi et al.~\cite{mohammadi2021improving} use two process-driven conceptual rainfall-runoff models—HBV (Hydrologiska Byrans Vattenbalansavdelning) and NRECA (Non Recorded Catchment Areas)—to provide inputs for support vector machines (SVMs) and an adaptive neuro-fuzzy inference system (ANFIS), resulting in seven hybrid model variants. The results indicate that AI-based hybrid models generally deliver more accurate streamflow estimates compared to the HBV and NRECA models alone. Moreover, three hybrid methods integrating HBV model simulations with LSTM~\cite{yu2023enhancing} are developed for semi-arid regions, demonstrating significant improvements over both the HBV and standalone LSTM models. Numerous studies illustrate the efficacy of serial hybrid methods in combining conceptual hydrological models with data-driven models.

\textbf{Parallel Hybrid.} In a parallel hybrid architecture, data-driven models and physics-driven models are integrated concurrently. This approach can employ a data-driven model to combine hydrological predictions in parallel. For instance, a hybrid approach combining conceptual and ML techniques has been proposed to enhance the accuracy of runoff simulations in snow-covered basins~\cite{achite2022enhancing}. An end-to-end hybrid modeling approach is introduced to learn and predict the spatiotemporal variations of observed and unobserved hydrological variables globally~\cite{kraft43hybrid}. This model integrates a dynamic neural network with a conceptual water balance model, constrained by water cycle observational products such as evapotranspiration, runoff, snow water equivalent, and changes in terrestrial water storage. The model accurately reproduces observed water cycle variations, and the relationships of runoff generation processes are well aligned with established understanding. Furthermore, a hybrid hydrological model~\cite{kraft2022towards} is used for global hydrological modeling, utilizing neural networks' adaptability to represent uncertain processes within a framework grounded in physical principles, such as mass conservation. This hybrid model is simultaneously trained using a multi-task learning approach, indicating that it provides a novel data-driven perspective for modeling the global hydrological cycle and physical responses through machine-learned parameters. This method aligns with existing global modeling frameworks. Moreover, parallel hybrid methods have been shown to correct model biases~\cite{watt2021correcting}. 

\blue{\textbf{Complex Hybrid (Coupling Serial and Parallel Hybrids).}  In a complex structure, data-driven models can be integrated into the input space, output space, and interior of hydrological models in a complex hybrid method, according to different application requirements. First, as shown in Fig.~\ref{fig:0}(c) top, data-driven models can parameterize hydrological models in the input space through serial hybrid. Simultaneously, data-driven models can effectively represent the internal dynamics of hydrological processes by integrating with traditional hydrological models.  For example, a general
physics‐AI approach~\cite{jiang2020improving} is proposed to improve AI geoscientific awareness, wherein temporal dynamic geoscientific models are included as a special process‐wrapped recurrent neural network (denoted as P‐RNN). Common NN layers that lack physics-based processes address those unrepresented by the P‐RNN layer. This combination improves the representation of rainfall-runoff processes. Additionally, a parameterization pipeline using NN layers through serial hybrid maps region-dependent attributes (e.g., soil properties and topography) onto the P-RNN and common NN layers.
Experiments with the physics-AI approach show that this complex hybrid model improves prediction accuracy, model transferability, and the ability to infer unobserved processes. Additionally, a differentiable programming framework~\cite{feng2022differentiable} is designed to parameterize, enhance, or replace the process-based model's modules in a simple hydrologic model, HBV, for predicting hydrologic variables such as streamflow.  Specifically, the LSTM outputs the physical parameters of the process-based model via serial hybrid method. Within the process-based model, certain components can be substituted with NNs and the structure is updatable. The framework is trained end-to-end, without intermediate supervising data or labels. The primary loss function is computed between the output of the process-based model and the observed data.
Second, as shown in Fig.~\ref{fig:0}(c) below, data-driven models not only effectively capture the internal dynamics of hydrological processes by integrating with traditional hydrological models but also enhance model prediction performance in the output space as a post-processing layer. For instance, a process-driven DL model~\cite{li2024process} is developed to enhance the process awareness of DL models. A conceptual hydrological model is integrated into an RNN cell to represent runoff sub-processes effectively, and an LSTM is used as a post-processor layer to output relevant system responses. Experimental results demonstrate that this complex hybrid model better characterizes the rainfall-runoff relationship, improving prediction performance.}

\section{HydroPML: Process-based Hydrology in Physics-aware Machine Learning \label{section3}}
\begin{figure}[!htp]
	\centering
	{\includegraphics[width = 0.9\textwidth]{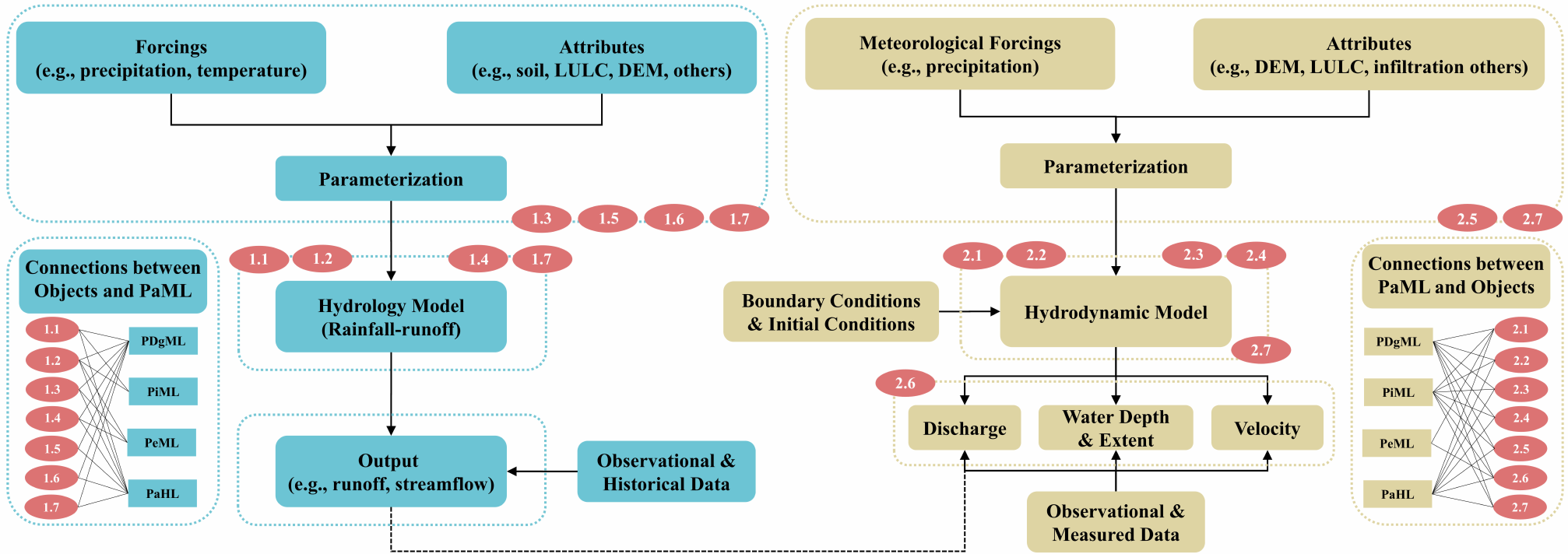}}
	\caption{\blue{The proposed process-based hydrology in physics-aware machine learning (HydroPML). HydroPML encompasses rainfall-runoff hydrological process understanding  over time scales ranging from hours to decades  and hydrodynamic process  understanding over time scales from seconds to days. 
 Red labels denote different objectives of HydroPML. For example, the oval labeled 1.1 represents short-term forecasts. The lower left and right corners represent the connections between PaML and the objects for rainfall-runoff hydrological processes and hydrodynamic processes, respectively.
		See the text for more detail.}}
	\label{fig:4}
\end{figure}

As shown in Fig.~\ref{fig:4}, the proposed HydroPML includes rainfall-runoff hydrological process understanding and hydrodynamic process  understanding.

\subsection{Rainfall-runoff Process Understanding}
Rainfall-runoff models are extensively utilized in hydrology to investigate hydrological processes and play a crucial role in water resources management, encompassing runoff prediction~\cite{hu2018deep}, flood prediction~\cite{masseroni2017reliable}, and drought analysis~\cite{peterson2021watersheds}. These models have been developed for diverse applications, ranging from small catchments to global scales. The hydrological model can be classified into three broad groups: the empirical model, conceptual models, and physical-based models~\cite{devia2015review}. 
Detailed information regarding Physical Methods in Hydrological Process is available in Appendix A of SM.

\subsubsection{Rainfall-runoff Forecast Meets Physics-aware Machine Learning}
The primary input to the rainfall-runoff models is past  forcing data (such as  precipitation, temperature), constant regional attributions (such as land use land cover (LULC), topography, and others) and the output is the future runoff or streamflow.
As shown in Table~\ref{tab:6}, improving process-based rainfall-runoff models requires progress on several fundamental research challenges:
(1) building appropriate methods to forecast the runoff for the different time scales within the model domain (Short-term Forecasts and Long-term Forecasts); (2) representing the variability of hydrologic processes across a hierarchy of spatial scales (Spatial Variability); (3) verifying model reliability (Bias Reduction and Model Reliability); (4) testing  a model
across the different model subdomains or ungauged basins (Missing Data and Ungauged Basins); (5) estimating input data and model parameters (Parameterization); and (6) characterizing model uncertainty (Uncertainty Estimation in Rainfall-runoff Forecast). 

\begin{table}[!htp]
	\caption{PaML-based rainfall-runoff hydrological forecasts classified by objectives.}
	\centering
	\resizebox{1.\textwidth}{!}
	{\Huge
		\begin{tabular}{p{8cm}p{15cm}p{15cm}p{15cm}p{15cm}}
		\hline
		\multicolumn{1}{c}{\textbf{Objectives}}                    & \multicolumn{1}{c}{\textbf{PDgML}}                                                                                                                                                                                                                                                                                                                                                                                                                                                                                                                                                                                                                                        & \multicolumn{1}{c}{\textbf{PiML}}                                                                                & \multicolumn{1}{c}{\textbf{PeML}}                                                                                                                                                                                                                                    & \multicolumn{1}{c}{\textbf{PaHL}}                                                                                                                                                                                                                                                                                                                                                            \\ \hline
		(1.1)   Short-term Forecasts                               & LSTM~\cite{hu2018deep, fu2020deep}, physics-guided LSTM~\cite{xie2021physics}, GRUs~\cite{xiang2020distributed}, graph convolutional GRUs~\cite{sit2021short}, graph convolutional RNN~\cite{zanfei2022graph}, stacked LSTM and bi-directional LSTM~\cite{wegayehu2022short}, ensemble model~\cite{granata2022stacked, di2023short}, ConvLSTM~\cite{chen2022short}, data-driven forecasting using only in-stream measurements~\cite{muste2022flood}
		 &  GW-PINN~\cite{zhang2022gw}
		                                                                  &                                                                                                                                                                                                                                                                      & Hybrid method combining hydrologic processes with LSTM~\cite{feng2020enhancing}, SWAT-LSTM~\cite{chen2023improving}, conceptual hydrological model by integrating NNs~\cite{kumanlioglu2019performance}, urban rainfall-runoff modeling~\cite{nareshurban},   Google’s operational flood forecasting system~\cite{nevo2022flood}  \\ \hline
		(1.2)   Long-term Forecasts                                & LSTM~\cite{kratzert2018rainfall}, Gaussian process regression~\cite{sun2014monthly}, deep belief network~\cite{yue2020mid}, extreme learning machine~\cite{alizadeh2021modeling},  
		an encoder-decoder CNN algorithm~\cite{herbert2021long}, Bayesian ensemble learning by combining different ML methods~\cite{he2023medium}
		                                                                                                                                                & LSTM by considering conservation of mass into loss function~\cite{khandelwal2020physics} & Entity-aware-LSTM~\cite{kratzert2019towards}, MC-LSTM~\cite{hoedt2021mc}, differentiable   modeling~\cite{shen2023differentiable}, P‐RNN layer~\cite{jiang2020improving} & A hybrid monthly streamflow forecasting approach~\cite{humphrey2016hybrid}, ML is embedded into a lumped conceptual rainfall-runoff model~\cite{okkan2021embedding}, three hybridization approaches~\cite{yu2023enhancing}, differentiable parameter learning~\cite{tsai2021calibration} \\ \hline
		(1.3)   Spatial Variability                                & Fully-distributed GNN~\cite{xiang2022fully}, graph convolutional GRUs based model~\cite{sit2021short}, graph convolutional RNN~\cite{zanfei2022graph}, ConvLSTM~\cite{chen2022short}, graph-based reinforcement learning~\cite{jia2021graph}, data synergy~\cite{fang2022data}, initializing a recurrent GNN by physical model~\cite{jia2021physics}                  &                                                                                                                  & HydroNets~\cite{moshe2020hydronets}, a differentiable, learnable physics-based routing model~\cite{bindas2022improving}
		                                                                                              & Hybrid approach combining a spatially-distributed snow model with convLSTM~\cite{xu2022hybrid}                                                                                                                               \\ \hline
		(1.4)   Bias Reduction and Reliability               & Bias-corrected remote sensing  products~\cite{zhu2023spatiotemporal}, directly measured datasets~\cite{muste2022flood},  ensemble methods~\cite{zounemat2021ensemble, liu2022ensemble}
                                                                                                 &                                                                                                                  & MC-LSTM~\cite{hoedt2021mc, frame2023strictly}, differentiable   modeling~\cite{shen2023differentiable}
      & LSTM using an additional input feature from the process-based model~\cite{wi2022assessing}, using physical models coupled with LSTM~\cite{liu2021observation}, WRF-Hydro-LSTM~\cite{cho2022improving}, LSTM daily streamflow prediction models~\cite{frame2021post}
       \\ \hline
		(1.5)   Missing Data and Ungauged Basins                   &    Attention-based deep learning (such as 3D-CNN-Transformer)~\cite{ghobadi2022improving}, input-selection ensemble method~\cite{feng2021mitigating}, \blue{encoder–decoder LSTMs~\cite{nearing2024global}, encoder-decoder double-layer LSTM~\cite{zhang2024deep}}                                                                                                                                                                                                                                                                                                                                                                                                                                                                                                                                                                                                                                                                    &                                                                                                                  &                                                                                                                                                                                                                                                                      & AI-based hybrid models~\cite{mohammadi2021improving}, serial hybrid LSTMs~\cite{konapala2020machine, lu2021streamflow}, transfer learning in SWAT-LSTM~\cite{chen2023improving}.
		                                                                                                                                                                                                                                                                                                                 \\ \hline
		(1.6)   Parameterization                                   & A machine learning approach~\cite{fang2022estimating}, deep neural network~\cite{jiang2022knowledge}, machine learning‐based maps~\cite{yang2019quest}                                                                                                                                                                                                                                                                                                                                                                                                                                                                                                                                                                                                                                                                          &                                                                                                                  &                                                                                                                                                                                                                                                                      & A differentiable neural network~\cite{feng2022differentiable}, differentiable parameter learning~\cite{tsai2021calibration}
  \\ \hline
		(1.7)   Uncertainty Estimation in Rainfall-runoff Forecast & Data-driven approach~\cite{pathiraja2018data}, Bayesian processor~\cite{liu2022uncertainty}, hybrid ensemble and variational data assimilation framework~\cite{abbaszadeh2019quest}, hierarchical model tree~\cite{solomatine2009novel}, probabilistic  deep ensembles~\cite{chaudhary2022flood}, LSTM-based approaches~\cite{zhu2021internal, althoff2021uncertainty}, uncertainty estimation benchmarking procedure and baselines~\cite{klotz2022uncertainty}, analysis framework for sources of uncertainties~\cite{song2020uncertainty}
            &                                                                                                                  &                                                                                                                                                                                                                                                                      & Using physical models coupled with LSTM~\cite{liu2021observation}, XAJ-MCQRNN~\cite{zhou2022short}, hybrid error correction model~\cite{roy2023novel}
        \\ \hline
	\end{tabular}}
\label{tab:6}
\end{table}

\textbf{Short-term Forecasts.} See Fig.~\ref{fig:4} (label 1.1).
Short-term forecasts focus on outlook horizons ranging from hours to weeks.
 Innovative forecasting tools are transforming flood early warning mechanisms, and agricultural and hydropower management schemes.
 PaML offers robust tools to address the formidable challenges associated with the uncertainties, data dependencies, and intricacies inherent in short-term hydrological forecasting. First, the significant uncertainty and complexity inherent in hydrological processes and weather-climate factors affecting river basins have increasingly motivated researchers to adopt PDgML approaches. Specifically,  the PDgML models based on LSTM~\cite{hu2018deep, fu2020deep} are proposed to predict streamflow. They have been found to have greater accuracy in predicting hourly and daily streamflow  than a classical NN with back-propagation.  A model~\cite{xiang2020distributed} utilizing multiple GRUs is proposed for predicting streamflow up to 120 hours ahead.
Granata et al.~\cite{granata2022stacked} demonstrate that their ensemble model, based on random forest and multilayer perceptron, outperforms bi-directional LSTM networks in predicting peak flow rates, while also achieving significantly shorter computation times.
 Wegayehu and Muluneh~\cite{wegayehu2022short} compare various ML algorithms for one-step daily streamflow forecasting, showing that both MLP and GRU models perform better than stacked LSTM and bi-directional LSTM models.
An ensemble model~\cite{di2023short} incorporating nonlinear autoregressive networks, multilayer perceptrons, and random forests is proposed for short-term streamflow forecasting, using precipitation as the only exogenous input, with a forecast horizon of up to 7 days.
In addition, to model the spatial diversity of rivers, the graph convolutional GRUs-based model~\cite{sit2021short} is designed to predict streamflow for the next 36 hours at a sensor location by utilizing data from the upstream river network.
Graph convolutional RNN~\cite{zanfei2022graph} is developed for robust water demand forecasting. 
More detailed information about PDgML studies on short-term  streamflow prediction can be found in~\cite{ibrahim2022review}.  Furthermore, to overcome data dependence, attempts have been made based on the PiML community, such as PINNs, for solving groundwater flow equations~\cite{zhang2022gw}. Because it is difficult to express the whole process of rainfall-runoff with specific equations, the current progress based on PiML technology is very slow. In addition, integrating hydrologic processes with LSTM~\cite{feng2020enhancing} proves effective in assimilating recent streamflow observations, thereby enhancing near-term daily streamflow forecasts. A coupled soil and water assessment tool (SWAT)-LSTM approach~\cite{chen2023improving} is developed  to provide  a potential shortcut for conducting daily streamflow simulations in both ungauged and poorly gauged watersheds.
The conceptual hydrological model by integrating NNs~\cite{kumanlioglu2019performance} reveals that the hybrid approach outperforms both the original GR4J model and the single NN-based runoff prediction model in terms of prediction accuracy.
 
 In general, existing conceptual and physical models are widely used in operational forecasting within the time scale of hours to weeks. The development of PaML-based short-term forecast methods is not very mature~\cite{slater2023hybrid}. However, PaML provides  promising alternatives to traditional physics-based hydrological and inundation models for flood prediction due to its automation, efficiency, and flexibility. For example,  the PDgML (ConvLSTM~\cite{chen2022short}, data-driven forecasting using only in-stream measurements~\cite{muste2022flood}) for extracting spatio-temporal features of hydrological information exceeds the recent models in predicting flood arrival time and peak discharge. PaHL-based methods (such as  Google’s operational flood forecasting system~\cite{nevo2022flood} and urban rainfall-runoff modeling~\cite{nareshurban}) show potential for flood nowcasting.

\textbf{Long-term Forecasts.} See Fig.~\ref{fig:4} (label 1.2). Long-term forecasts focus on outlook horizons of sub-seasonal, year, to decade. The vast majority of ML predictions in the literature are PDgML methods, which are driven by observational or historical data. PDgML captures intricate non-linear patterns from data that would otherwise necessitate extrapolation, offering an alternative approach for long-term rainfall-runoff predictions, for example,  LSTM~\cite{kratzert2018rainfall}, Gaussian process regression~\cite{sun2014monthly}, deep belief network~\cite{yue2020mid}, extreme learning machine~\cite{alizadeh2021modeling},  
an encoder-decoder CNN algorithm~\cite{herbert2021long}, and Bayesian ensemble learning by combining different ML methods~\cite{he2023medium}. These approaches demonstrate high accuracy in analyzing the relationship between rainfall and runoff, as well as the interaction between climate factors and rainfall. 
In order to alleviate the dependence on data, a physics-guided LSTM~\cite{khandelwal2020physics} that considers the conservation of mass into loss function  is proposed. Digging deeper into physics (such as mass conservation and similarities) in the hydrological process, MC-LSTM~\cite{hoedt2021mc}  is introduced by modifying the LSTM to ensure mass input conservation. Entity-Aware-LSTM~\cite{kratzert2019towards} learns catchment similarities as a feature layer within a DL model. 
In addition, based on the PeML technology, some NN modules with physical properties (such as differentiable   modeling~\cite{shen2023differentiable} and the P‐RNN layer in the general
physics‐AI approach~\cite{jiang2020improving}) are proposed for rainfall-runoff forecasts. Furthermore, hybrid learning models (PaHL) can be suitable alternatives to hydrological models. Hybrid learning based on conceptual hydrological models tries to combine the strengths of  hydrological models and data-driven methods. For example, a hybrid monthly streamflow forecasting approach~\cite{humphrey2016hybrid} and three hybridization techniques~\cite{yu2023enhancing} demonstrate that the hybrid model not only significantly surpasses conceptual hydrological models in precision but also outperforms purely ML models. 

However, in many hybrid modeling methods, variables generated by the conceptual model are utilized as inputs for a data-driven model, which generally leads to increased computation time. Additionally, these interactions are often overlooked because the two models are calibrated separately.
Thus, embedded ML in a lumped conceptual rainfall-runoff model~\cite{okkan2021embedding} is proposed by combining the two models and conducting their calibration jointly, and differentiable parameter learning~\cite{tsai2021calibration} is presented to  efficiently learn a hybrid model in an end-to-end manner.

\textbf{Spatial Variability.} See Fig.~\ref{fig:4} (label 1.3). Rainfall-runoff modeling is difficult due to the  non-linear structure of the process, which involves much spatio-temporal variability.
Spatial variability is critical for predictions in ungauged basins and for developing universal regional models. PaML (especially GNNs) provides the possibility of distributed modeling. Specifically, for PDgML, GNN~\cite{xiang2022fully} is used for fully-distributed rainfall-runoff modeling. The result shows that  the GNN-based fully-distributed model~\cite{xiang2022fully} successfully represents the spatial information in predictions. A  graph convolutional GRUs based model~\cite{sit2021short}, ConvLSTM~\cite{chen2022short}, and graph convolutional RNN~\cite{zanfei2022graph} can also be developed to model spatial diversity of rivers. 
A recurrent GNN~\cite{jia2021physics} is proposed to capture the interactions
among multiple segments in the river network. In this work,
a physics-based model is used to initialize the ML model and learn
the physics of streamflow.
To mitigate the requirement for large training datasets, a real-time active learning approach is employed in a graph-based reinforcement learning framework~\cite{jia2021graph}. This method leverages spatial and temporal contextual information to select representative query samples.
In addition, region-specific differences can be learned for the LSTM model by inputting sufficient heterogeneous data (data synergy)~\cite{fang2022data}. 

Furthermore, spatial physical characteristics of rivers (such as river networks and hydrologic routing modules) can be encoded into NNs based on the PeML technology. For example, a hydrologic model called HydroNets~\cite{moshe2020hydronets} is built by leveraging river network structure. Specifically, each node in the graph  represents a basin, and an edge direction corresponds to water flow from a sub-basin to its containing basin. A differentiable, learnable physics-based routing model~\cite{bindas2022improving} is proposed to  constrain spatially-distributed parameterization. This marks the first instance of an interpretable, physics-based model being developed on the river network to infer spatially-distributed parameters. Finally, for the PaHL, the hybrid models combining conceptual models and data-driven models  are often lumped, i.e., spatially averaged, at the catchment scale. To address this problem, a hybrid approach~\cite{xu2022hybrid} is developed to predict streamflow. The hybrid approach combines a convLSTM with a spatially-distributed snow model to simulate the effect of surface and subsurface properties on streamflow. In the future, building a robust spatio-temporal representation for different watersheds based on the PaML community will be an ongoing direction for research.

\textbf{Bias Reduction and Reliability.} See Fig.~\ref{fig:4} (label 1.4).
One important challenge of PaML-based rainfall-runoff forecasts is the ability of such models to produce reliable or physically-plausible results in different domains, especially extreme conditions such as floods and droughts. Specifically, reliable spatio-temporal  remote sensing data and directly measured datasets can be used to reduce the bias of  PDgML-based rainfall-runoff forecasting. For example, bias-corrected remote-sensing precipitation products  are adopted  as precipitation input  of a spatio-temporal DL rainfall-runoff forecasting model~\cite{zhu2023spatiotemporal}, which can improve the model's reliability for extreme flood forecasts. Directly measured datasets are utilized for streamflow estimation to develop a data-driven forecasting algorithm~\cite{muste2022flood}.
However, these bias-corrected data-based approaches suffer from deficiencies over small catchments due to uncertainty in remote sensing data and errors in hydrological models. Thus, some ensemble methods (such  as resampling ensemble methods~\cite{zounemat2021ensemble}, which  combine hydrometeorological modeling and LSTM~\cite{liu2022ensemble}) are explored to improve the reliability and robustness of PDgML models.
Furthermore, 
one emerging route for bias reduction and reliability is to employ PeML designs that explicitly observe the physical laws in the rainfall-runoff process, such as  MC-LSTM~\cite{hoedt2021mc} and differentiable   modeling~\cite{shen2023differentiable}.
Another new development is hybrid models to produce physically-plausible or explainable results. For instance, Wi et al.~\cite{wi2022assessing} explore the capability of DL models to provide reliable future projections of streamflow under warming conditions. It concludes that  integrating estimates of evapotranspiration from process-based models as additional input features can enhance the performance of LSTM models in hydrologic projections under climate change. Alternatively, training LSTMs on a diverse set of watersheds also proves effective. 
Another approach involves coupling physical models (such as global climate models or global hydrological models) with LSTM models~\cite{liu2021observation} for long-term streamflow projections. Here, the LSTM acts as a post-processor aimed at constraining streamflow simulations derived from physics-based models to reduce uncertainty.
 To improve streamflow prediction within the Weather Research and Forecasting Hydrological Modeling System (WRF-Hydro), a method known as WRF-Hydro-LSTM~\cite{cho2022improving} is proposed. This method utilizes LSTM models to predict residual errors within WRF-Hydro's simulations.

However, due to data errors in hydrological records,  an unconstrained ML (such as LSTM) outperforms a PeML model (such as MC-LSTM) because of the ability of PDgML to learn and accommodate these data errors~\cite{frame2023strictly}. In addition, three  LSTM-based daily streamflow prediction models~\cite{frame2021post} are built and compared, including a LSTM post-processor trained on outputs from the United States National Water Model (NWM), a LSTM post-processor trained on NWM outputs and atmospheric forcings, and a LSTM model trained solely on atmospheric forcings.  It concludes that a LSTM model trained only on atmospheric forcing outperforms the other two LSTMs in ungauged basins. Hybrid models may reduce the reliability of the model in some cases. Thus, PiML, PeML, and PaHL approaches for the reliability of rainfall-runoff forecasts require further development in the future.  

\textbf{Missing Data and Ungauged Basins.} See Fig.~\ref{fig:4} (label 1.5). Hydrologic predictions at many watersheds, such as rural watersheds, as well as cold and ungauged regions, are important but also challenging due to data scarcity~\cite{belvederesi2022modelling}. Thus, many PaML-based methods are developed for poorly gauged or  ungauged basins. For example, geo-spatio-temporal mesoscale data and attention-based deep learning~\cite{ghobadi2022improving} are used to improve long-term streamflow prediction. This study demonstrates the superior performance of the 3D-CNN–Transformer compared to TD-CNN–Transformer, TD-CNN–LSTNet, and 3D-CNN–LSTNet in hydroclimate data applications, particularly for poorly gauged basins. Despite the excellent performance of deep NNs in streamflow predictions, their effectiveness declines in ungauged regions. To mitigate these errors, an input-selection ensemble method~\cite{feng2021mitigating} is proposed, leveraging the flexibility of deep networks to integrate satellite-based soil moisture products or daily flow distributions. \blue{Recently, a watershed-aware streamflow forecast model, an encoder-decoder double-layer LSTM~\cite{zhang2024deep}, is proposed by encoding basin spatial properties into LSTM. This model demonstrates its effectiveness in data-scarce regions such as Chile. Concurrently, another model, encoder–decoder LSTMs~\cite{nearing2024global},  is developed for streamflow forecasts at a global scale. This encoder–decoder model reliably predicts extreme river events with a lead time of up to 5 days in ungauged basins. These advancements highlight the significant potential of PDgML for streamflow and flood forecasting in ungauged basins.}
Furthermore, hybrid models present viable alternatives to hydrological models, especially in watersheds lacking measured data such as climatic parameters, land cover, soil data, and groundwater aquifer properties, provided that appropriate inputs like rainfall are available. Mohammadi et al.~\cite{mohammadi2021improving} demonstrate that AI-based hybrid models generally yield more accurate streamflow estimates than conceptual rainfall-runoff models in ungauged basins. \blue{Additionally, the hybrid framework~\cite{mohammadi2024conceptual} combines a conceptual hydrological model (such as Glacial Snow Melt (GSM)) with Support Vector Regression (SVR) and firefly algorithm-driven parameter optimization, demonstrating superior performance in rainfall runoff prediction compared to standalone GSM and conventional SVR.}
To improve out-of-distribution prediction in data-scarce basins, several sequential methods have been proposed for developing hybrid LSTMs~\cite{konapala2020machine, lu2021streamflow}. These methods involve using outputs from physics-based hydrologic model simulations as additional inputs in LSTM networks. Additionally, to model long-duration daily streamflow in poorly gauged watersheds, transfer learning techniques are employed with a pre-trained SWAT-LSTM model~\cite{chen2023improving}.

However, these PaML-based approaches are still only preliminary attempts. Deeply exploiting the transferability of ML models (such as domain adaptation methods~\cite{nearing2023hindcast,xu2023universal},  meta-learning methods~\cite{chen2023physics}) and the interpretability of physical models will effectively solve hydrological prediction problems in ungauged basins. Furthermore, leveraging the increasing volume of simulation and observation data in hydrology, we can integrate large pre-trained models (e.g., generative models in ChatGPT~\cite{foroumandichatgpt}, Transformer in ClimaX~\cite{nguyen2023climax}) with hydrological knowledge to establish a foundational hydrology pre-training model. The pre-training large model can then be finely tuned to address diverse hydrological prediction tasks, including those in ungauged basins. 


\textbf{Parameterization.} See Fig.~\ref{fig:4} (label 1.6). Parameters for predicting rainfall-runoff are typically not directly measurable; instead, they are indirectly estimated through prior knowledge or model calibration. Calibration  involves adjusting model parameters to match outputs with observed data at specific locations.  Specifically, a ML approach and remotely sensed underlying surface data~\cite{fang2022estimating}  are employed to develop models for estimating the runoff routing parameter. This study demonstrates that ML-derived parameters from extensive datasets exhibit robustness, ensuring the overall performance of physical models. A deep neural network~\cite{jiang2022knowledge} is used to  construct inverse mapping from informative responses to each of the selected parameters. Yang et al.~\cite{yang2019quest} calibrate three essential parameters of the Variable Infiltration Capacity  model at every 1/8  grid‐cell employing ML‐based maps (shuffled complex evolution  algorithm). Furthermore, some PaHL-based methods, such as a differentiable programming framework~\cite{feng2022differentiable} and differentiable parameter learning~\cite{tsai2021calibration}, are proposed to    efficiently learn a  mapping between inputs  and
parameters of process-based models in an end-to-end manner. 

Parameterization is an important process of rainfall-runoff forecasts.  Digging deeper into the physical properties of hydrological processes (such as mass conservation), we can achieve fast and effective parameterization based on the PiML, PeML, and PaHL technologies.

\textbf{Uncertainty Estimation in Rainfall-runoff Forecasts.} See Fig.~\ref{fig:4} (label 1.7). Hydrological forecasts often suffer from inaccuracies due to insufficient conceptualization of underlying physics, inaccurate predictions of ML models, non-uniqueness of model parameters, and uncertainties in calibration data (such as streamflow outputs). Therefore, precise uncertainty estimations are crucial for actionable hydrological predictions. However, the majority of PaML-based rainfall-runoff studies do not provide uncertainty estimates.   In order to address this problem, some PDgML approaches are proposed  for model uncertainty, such as a data-driven approach~\cite{pathiraja2018data}, Bayesian processor~\cite{liu2022uncertainty},  a hybrid ensemble and variational data assimilation framework~\cite{abbaszadeh2019quest}, a hierarchical model tree~\cite{solomatine2009novel}, and probabilistic  deep ensembles~\cite{chaudhary2022flood}. In addition, several LSTM-based approaches are presented for uncertainty estimation. Two methods~\cite{zhu2021internal} are explored for integrating LSTM with Gaussian processes. In the first method, the LSTM is used to parameterize a Gaussian process, and in the second, the LSTM acts as a forecast model with a Gaussian process post-processor. Althoff et al.~\cite{althoff2021uncertainty} investigate predictive uncertainty in hydrological models based on LSTM by comparing multiparameter ensembles with dropout ensembles. Song et al.~\cite{song2020uncertainty} propose a framework or quantifying uncertainty contributions from the sample set, ML method, ML architecture, and their interactions in multi-step time-series forecasting using variance analysis.
Furthermore, to reduce the gap between the hydrological sciences and  ML  community for uncertainty estimations, Klotz et al.~\cite{klotz2022uncertainty} establish an uncertainty estimation benchmarking procedure and present four PDgML (LSTM) baselines, including three based on mixture density networks and one based on Monte Carlo dropout.
In addition, by integrating physics and ML approaches, PaHL has also been gaining attention,  as it can alleviate the computational burden and improve efficiency for uncertainty estimation. For example, the LSTM model is considered a post-processor~\cite{liu2021observation} that aims to  reduce the uncertainty of streamflow simulations from the physics-based model. Serial hybrid LSTM~\cite{lu2021streamflow} is introduced for  uncertainty quantification. An integrated model (XAJ-MCQRNN)~\cite{zhou2022short} is developed, incorporating the Xinanjiang conceptual model and composite quantile regression NN, to address error propagation and accumulation in multi-step flood probability density forecasts. Additionally, a hybrid error correction model~\cite{roy2023novel} is proposed to enhance streamflow forecast accuracy. The hybrid model integrates the HBV model with the ML algorithm through a data assimilation technique. 

In general, integrating physics knowledge into ML for uncertainty estimation has the potential to better characterize uncertainty. For instance, ML surrogate models might yield physically inconsistent predictions; integrating physical principles can help alleviate this problem. \blue{Additionally, 
ML can reduce the uncertainty of rainfall-runoff models by effectively extracting features and identifying significant variables~\cite{mohammadi2022ihacres}.
The reduced data needs of ML due to constraints for adherence to known rainfall-runoff processes could alleviate some of the computational costs associated with NNs. However, rainfall-runoff modeling remains a complex process, and defining a simple, unified benchmark—encompassing effective and publicly available datasets, robust metrics, and comprehensive baseline tools—for different PaML methods is a significant challenge~\cite{klotz2022uncertainty}. Furthermore, current PaML-based methods mainly focus on PDgML and PaHL, which face substantial challenges in addressing effectiveness, computational complexity, and usability in uncertainty estimation for rainfall-runoff forecasts. Therefore, deepening the integration of physical information from the rainfall-runoff process into ML frameworks or optimization processes (such as PeML and PiML) to achieve low uncertainty, end-to-end, and effective rainfall-runoff process modeling will be a crucial direction for future research.}

\subsection{Hydrodynamic Process Understanding}
Hydrodynamic models are mathematical models that describes fluid movement and typically require numerical methods for accurate solutions. These models simulate water motion by solving equations derived from the application of fundamental laws of physics, encompassing principles governing the conservation of mass, momentum, and energy. Hydrodynamic models can realize the simulation of hydrological processes (such as floods and rainfall-induced landslides) in a time scale of seconds, hours, or days. Hydrodynamic models are classified into one-dimensional (1D), two-dimensional (2D), and three-dimensional (3D) models based on the spatial representation of the flow. Detailed information about Physical Methods in Hydrodynamic Modeling is provided in Appendix B of SM.

\subsubsection{Hydrodynamic Modeling Meets Physics-aware Machine Learning}
This section provides a brief overview of a diverse set of objectives where hydrodynamic modeling meets PaML. As shown in Table~\ref{Tab:5}, improving hydrodynamic models requires progress on several fundamental research challenges, including solving hydrodynamic equations, scalability,  improving model generalizability and transferability, speed and operability, parameterization and calibration, data generation, and uncertainty quantification in hydrodynamic modeling.  These challenges are discussed in more detail below.

\begin{table}[!htp]
	\caption{PaML-based hydrodynamic modeling classified by objectives.}
	\centering
	\resizebox{1.0\textwidth}{!}
	{\Huge
	\begin{tabular}{p{8cm}p{18cm}p{18cm}p{12cm}p{12cm}}
		\hline
		\multicolumn{1}{c}{\textbf{Objectives}}                     & \multicolumn{1}{c}{\textbf{PDgML}}                                                                                                                                                                                                                                                                                                                                                                                                                                                                                               & \multicolumn{1}{c}{\textbf{PiML}}                                                                                                                                                                                                                                                                                                                                                                                                                     & \multicolumn{1}{c}{\textbf{PeML}}                                                                                                                                                                                                                                                                                                                                                                                          & \multicolumn{1}{c}{\textbf{PaHL}}                                                                                                                                                                                                     \\ \hline
		(2.1)   Solving Hydrodynamic Equations                      & Flood inundation modeling~\cite{karim2023review}, coastal bridge hydrodynamics~\cite{xu2023machine}, CNN models~\cite{kabir2020deep, guo2021data}, data-driven models~\cite{zheng2023data}, DCGAN~\cite{cheng2021real}, DRL-based method~\cite{feng2023control}, ANFIS~\cite{zhang2022three}, HIGNN~\cite{ma2022fast}
		 & Flow physics-informed learning~\cite{cai2021physics}, PINNs for spatial-temporal flood forecasting~\cite{mahesh2022physics}, PINNs for river channel~\cite{nazari2022physics}, PINNs for solving 2D shallow-water equations~\cite{bihlo2022physics,li2023physical}, NSFnets~\cite{jin2021nsfnets}, PINNs for 1D flood routing~\cite{bojovic2022physics} 
		  & FGN~\cite{li2022graph} and continuous convolutions~\cite{ummenhofer2020lagrangian} for Lagrangian fluid simulation of Navier-Stokes equations, differentiable modeling~\cite{shen2023differentiable}, SCG-NN~\cite{rahman2021development}, GRU-HD~\cite{huang2022coupling}
		   & Differentiable hybrid approach~\cite{nava2022fast}, deep learning-based shallow water equations solver~\cite{forghani2021deep}, GRU-HD~\cite{huang2022coupling} \\ \hline
		(2.2)   Scalability                                         & Data-driven discretization~\cite{bar2019learning}, data-driven subgrid approach~\cite{ye2021data}                                                                                                                                                                                                                                                                                                                                                                  &  PINN of  the Saint-Venant equations~\cite{feng2023physics}, GeoPINS  of 2D shallow water equations~\cite{xu2023large}                                                                                                                                                                                                                                                                                 &                                                                                                                                                                                                                                                                                                                                                                                                                            & Physics-aware downsampling method~\cite{giladi2021physics}
		                                                                                                                                                  \\ \hline
		(2.3)   Model Gneralizability and Transferrability          & Transfer learning~\cite{seleem2023transferability}, data-driven forecasting based on dynamical systems theory~\cite{okuno2021practical}, transfer learning enhanced DeepONet~\cite{xu2022transfer}
		                                                                                                                                                                                                                                                      & Meta-learning based PINNs~\cite{de2021hyperpinn, psaros2022meta, penwarden2021physics}, TPINN~\cite{manikkan2022transfer}                                                                                                                                                                                                                                                            &                                                                                                                                                                                                                                                                                                                                                                                & Geometry transferability of HINTS~\cite{kahana2023geometry}, hybrid hydrodynamic models for accelerating 2D flood models~\cite{jamali2021machine}                                     \\ \hline
		(2.4)   Speed and Operability                               & DCGAN~\cite{cheng2021real}, random forest~\cite{zahura2020training}                                                                                                                                                                                                                                                                                                                                                                                                                &  PINN of  the Saint-Venant equations~\cite{feng2023physics}                                                                                                                                                                                                                                                                                                                                                                    &                                                                                                                                                                                                                                                                                                                                                                                                                            & Hybrid framework combining the physics of fluid motion with probabilistic methods~\cite{ivanov2021breaking}, Google’s operational flood forecasting system~\cite{nevo2022flood}
		       \\ \hline
		(2.5)   Parameterization and Calibration                    & Hybrid genetic-instance based learning algorithm~\cite{ostfeld2005hybrid}, SLDA~\cite{wei4069683statistical}                                                                                                                                                                                                                                                                                                                                                             & Active training of PINNs~~\cite{arthurs2021active}                                                                                                                                                                                                                                                                                                                                                                      &                                                                                                                                                                                                                                                                                                                                                                                                                            & Differentiable parameter learning~\cite{tsai2021calibration}                                                                                                                                                  \\ \hline
		(2.6)   Data Generation                                     & Integrated hydrodynamic and ML models~\cite{sampurno2022integrated}, FloodGAN~\cite{hofmann2021floodgan}, conditional GAN~\cite{farimani2017deep}                                                                                                                                                                                                                                                                                            & Enforcing conservation laws in GAN~\cite{yang2019enforcing}, physics-informed GAN~\cite{meng2022learning}
		                                                                                                                                                                                                                                                                                               & Encoding invariances in GAN~\cite{shah2019encoding, joshi2020invnet}
                                                                                                                                                                                                                     & Hybrid approach for flood susceptibility assessment~\cite{fang2022hybrid}                                                                                                                                    \\ \hline
		(2.7) Uncertainty Quantification in Hydrodynamic Modeling & Object-based correction using MLobject-based correction~\cite{cooper2019object}, FABDEM~\cite{hawker202230},  random forest regression~\cite{gang2022flood}, Markov chain Monte Carlo approach~\cite{siripatana2017assessing}, Bayesian calibration~\cite{schwindt2023bayesian}                                                                                    & Physics-informed Gaussian process regression~\cite{kohanpur2023urban}, adversarial uncertainty quantification in PINNs~\cite{yang2019adversarial}, Bayesian PINNs~\cite{yang2021b}                                                                                                                                                                                        &                                                                                                                                                                                                                                                                                                                                                                                                                            & Hybrid frameworks~\cite{hosseiny2020framework}                                                                                                                      \\ \hline
	\end{tabular}}
\label{Tab:5}
\end{table}

\textbf{Solving Hydrodynamic Equations.} See Fig.~\ref{fig:4} (label 2.1). PaML is widely used to solve 1D, 2D, and 3D hydrodynamic equations, including PDgML, PiML, PeML, and PaHL. 

First, PDgML shows great potential for real-time hydrodynamic modeling/forecasting due to its simplicity, superior performance, and computational efficiency. 
Data-driven models~\cite{zheng2023data,karim2023review,xu2023machine}  provide a new phenomenological approach to learning constitutive relations of hydrodynamics from data.  Specifically,  CNN techniques~\cite{kabir2020deep, guo2021data} are applied in 2D hydraulic/hydrodynamic models to predict water depths using outputs from existing hydraulic models. Deep convolutional generative adversarial networks (DCGANs)~\cite{cheng2021real} have been developed for real-time flow forecasting.
For real-time feedback control of 2D hydrodynamics, a DRL-based feedback strategy~\cite{feng2023control} is proposed, focusing on controlling the 2D hydrodynamic on fluidic pinball, i.e., force extremum and tracking, from cylinders’ rotation. An adaptive neural fuzzy inference system (ANFIS)~\cite{zhang2022three} is utilized for modeling 3D flows in large-scale rivers. The learnable coefficients of ANFIS are trained on 3D flood flow dynamics data using large-eddy simulation.
In addition, for a Lagrangian form of hydrodynamics,  a hydrodynamic interaction graph neural network (HIGNN)~\cite{ma2022fast} is introduced for inferring and predicting particle dynamics in Stokes suspensions.
A second use of PaML can be found in PiML. Cai et al.~\cite{cai2021physics} review physics-informed learning of flow dynamics, which integrates data and mathematical models through the use of PINNs. Specifically, PINNs of the Saint-Venant equations~\cite{mahesh2022physics, bojovic2022physics} are developed for spatial-temporal scale flood forecasting. A  PINN built directly from a configuration of the Saint-Venant equations~\cite{nazari2022physics} is created for use in real river channels, demonstrating high prediction accuracy and scientifically consistent behavior.  PINNs~\cite{bihlo2022physics,li2023physical} are also utilized for solving 2D shallow-water equations. NSFnets~\cite{jin2021nsfnets} is developed for the incompressible Navier-Stokes equations.
Most recently, PINNs (PiML) is widely used in the field of hydrodynamics~\cite{dai2023physics}. However, the current generation of PINNs lacks the accuracy and efficiency of high-order CFD codes for solving hydrodynamic equations. State-of-the-art PiML methods, including advanced PINNs for addressing the challenges of vanilla PINNs, and the PINOs in Supplementary Table 2, can be used to alleviate the gap between PiML and hydrodynamic equation solutions.
Third, for PeML, these hydrodynamic equations (such as Lagrangian forms of Navier-Stokes equations, and differential equations) are used to design the frameworks or modules of NNs, such as FGN~\cite{li2022graph}, continuous convolutions~\cite{ummenhofer2020lagrangian}, and differentiable modeling~\cite{shen2023differentiable}. In addition, the hydrodynamic model–FLO 2D, coupled with a ML algorithm-scaled conjugate gradient neural network (SCG-NN)~\cite{rahman2021development}, is developed by embedding
flood properties (such as rainfall intensity, drainage density, soil type, land cover, and others) into the input spaces of NNs. The hydrological boundaries (such as the lake boundary and downstream boundary) can be embedded into the gated recurrent unit with a 1D HydroDynamic model (GRU‐HD)~\cite{huang2022coupling}. However, these PeML methods still lack interpretability. The deep embedding of data-driven methods and physics-driven methods is urgently needed. 
Finally, PaHL is a more flexible means of solving hydrodynamic equations. For example, parallel hybrid methods (such as the differentiable hybrid approach~\cite{nava2022fast}, and deep learning-based shallow water equations solver~\cite{forghani2021deep}) are designed to capture the hydrodynamic process of the fluid. Parallel hybrid methods (such as GRU‐HD~\cite{huang2022coupling}) are proposed to quickly and accurately simulate the water process. 

All of these methods for solving hydrodynamic equations are still preliminary attempts in the PaML community. Compared with physics-based hydrodynamic methods, HydroPML-based methods for solving hydrodynamic equations lack transferability and generalization across different domains. In addition, the absence of benchmark data to assess model performance poses a significant constraint on the development of efficient PaML models for hydrodynamic modeling.  It is envisioned that research in hydrodynamics could
be further advanced with the evolving PaML technologies.

\textbf{Scalability.} See Fig.~\ref{fig:4} (label 2.2). Hydrodynamic methods are highly dependent on resolution. The scalability of hydrodynamic methods means that coarse grids are fast but low-accurate, and fine grids are accurate but slow.  Given the advantages of PaML (such as resolution invariance in FNOs, mesh-free in PiML, and others), the scale effect of resolution can be balanced. For example, a physics-aware downsampling method~\cite{giladi2021physics} for scalable 2D hydrodynamic modeling  is proposed  by minimizing the distance between the solutions on the fine and coarse grids. 
Furthermore, PINN is mesh-free, making it an efficient tool for downscaling. Thus, PINN of  the Saint-Venant equations~\cite{feng2023physics} is proposed to simulate the downscaled flow for a large-scale river model.  PINN-based downscaling leverages observational data assimilation to generate more accurate subgrid solutions for a long-channel water depth. Based on the resolution invariance in FNOs, GeoPINS  of 2D shallow water equations~\cite{xu2023large} is proposed for large-scale flood modeling by training a robust deep neural operator on the coarse grid and then predicting directly on the fine grid. In addition, a host of data-driven methods (PDgML) are developed for learning optimized approximations to hydrodynamic equations based on actual solutions to the known underlying equations. For example, data-driven discretization~\cite{bar2019learning}  allows for the integration of a set of nonlinear equations in spatial dimensions at resolutions 4 to 8 times coarser than standard finite-difference methods. A data-driven approach~\cite{ye2021data} integrates subgrid-scale geometrical details into a regular coarse grid for modeling coastal hydrodynamics, enabling accurate solutions even with a large subgrid ratio. 

However, the application of downscaling in HydroPML to
large-scale modeling is still limited because (a) large-scale hydrodynamic models have a complex terrain surface, land use, and land cover and depend on a large number of model parameters; and (b) to date, there is no consistent PaML method to integrate both remote sensing data and in-situ measurements.

\textbf{Model Generalizability and Transferability.}  See Fig.~\ref{fig:4} (label 2.3). Despite the high accuracy of physics-based hydrodynamic methods, these models have long simulation run times
and therefore are of limited use for exploratory or real-time flood predictions. A robust and efficient PaML can replace the computationally expensive parts of hydrodynamic models  by improving the  generalizability and transferability of PaML models. Specifically, the NN models' transferability  for predicting water depth~\cite{seleem2023transferability}  can be further enhanced through the utilization of transfer learning techniques. A transfer learning technique is also utilized to successively correct the trained DeepONet at the prediction steps~\cite{xu2022transfer}. 
A practical data-driven forecasting method based on dynamic systems theory~\cite{okuno2021practical} is developed to predict unprecedented water levels. For PiML, there are many derivatives of PINNs designed to enhance generalizability and transferability for solving PDEs, such as meta-learning based PINNs~\cite{de2021hyperpinn, psaros2022meta, penwarden2021physics} and TPINN~\cite{manikkan2022transfer}. In addition, in order to improve the generalizability and transferability of PaHL methods, Jamali et al.~\cite{jamali2021machine} replace the time-consuming part of the physics-based
numerical models with more efficient data-driven approaches, increasing their computational efficiency and achieving good generalization capability.
Kahana et al.~\cite{kahana2023geometry} explore the geometry transferability properties of HINTS~\cite{zhang2022hybrid}. 

However,  these methods cannot be successful in real case studies (especially for large-scale hydrodynamic problems).  Future research should test the performance of PaML in terms of different space-time domains and the approximation of complex and practical hydrodynamic equations.

\textbf{Speed and Operability.} See Fig.~\ref{fig:4} (label 2.4). Real-time and operational hydrodynamic modeling (such as flood forecasting) is crucial for supporting emergency responses and reducing risk and damage~\cite{hallegatte2012cost}.  Currently, there are few operational systems that forecast flooding at spatial resolutions that can facilitate emergency preparedness and response actions to mitigate flood impacts. Combining the fast inference of ML models with the reliability of physical models, real-time and operability of hydrodynamic modeling is possible. Specifically, a real-time predictive DCGAN~\cite{cheng2021real} is developed for forecasting floods. The last one-time step-ahead forecast from the DCGAN can serve as a new input for the subsequent time step-ahead forecast, which recursively forms a long lead-time forecast. Random forest~\cite{zahura2020training} serves as a surrogate model for real‐time flood prediction at a street scale. The surrogate model is trained using hourly water depths simulated by a 1D/2D hydrodynamic model. In addition, this PiML-based method~\cite{feng2023physics}  for reducing the computational load of traditional hydrodynamic methods (such as scalable hydrodynamic modeling) can be used for real-time hydrodynamic modeling. Furthermore, a hybrid framework for real-time flood modeling that merges fluid dynamics with probabilistic techniques~\cite{ivanov2021breaking} is developed to address the excessive computational requirements of high-fidelity real-time modeling. Google's operational flood forecasting system~\cite{nevo2022flood}, which integrates hybrid hydrological and hydrodynamic models, is established to deliver precise real-time flood alerts to agencies and the public. This forecasting system includes data validation, stage prediction, inundation modeling, and alert dissemination.

However, a major limitation of speed and operability is the lack of observational and  measured  data. In order to address this problem, some advanced PiML technologies for incomplete, noisy input data, as well as PeML and PaHL for efficient use of few-shot or unsupervised learning schemes, are areas that can be explored in the future.

\textbf{Parameterization and Calibration.} See Fig.~\ref{fig:4} (label 2.5). The behaviors of hydrodynamic models depend heavily on parameters (such as river bed elevation, channel geometry, precipitation, land use land cover, etc.) that need calibration. However, traditional calibration (such as manually-calibrated process-based models) is highly inefficient and results in nonunique solutions~\cite{tsai2021calibration}. Thus, many PaML-based parameter learning methods are proposed for hydrodynamic modeling. Specifically,  
a genetic-k-nearest neighbor hybrid algorithm~\cite{ostfeld2005hybrid} is proposed to mitigate the impractically high computational efforts associated with conventional calibration search techniques, while still achieving high-quality calibration results.
A set of statistical-learning-based data assimilation (SLDA) methods~\cite{wei4069683statistical} is proposed for estimating parameters in hydrodynamic models. In addition, according to the PiML community,  PINNs can be used to  solve parameter identification and parameter learning (data-physics-driven parameter discovery). For example, PINNs~\cite{arthurs2021active} are actively trained to approximate solutions to the Navier-Stokes equations over a parameter space region, where these parameters define physical properties like domain shape and boundary conditions.
 Furthermore, differentiable parameter learning~\cite{tsai2021calibration} is proposed to  efficiently learn a global mapping between inputs (and optionally responses) and
parameters.  Differentiable parameter learning contains a parameter estimation module that maps from raw input information  to process-based model parameters, as well as a differentiable  process-based model. The use of differentiable parameter learning leads to improved performance, enhanced physical coherence, superior generalizability, and reduced computational cost. 

However, these parameter learning methods basically perform supervised parameter correction based on limited observation data or measured data. These methods may lack generalization and depend on reliable observation data. In the future, physics-discovery NNs, data-physics-driven parameter discovery, or PaHL (such as differentiable parameter learning~\cite{tsai2021calibration}) may become an effective tool to solve these problems.

\textbf{Data Generation.} See Fig.~\ref{fig:4} (label 2.6). Data generation methods are valuable for simulating hydrodynamic processes under specific conditions. For instance, an integrated model~\cite{sampurno2022integrated} combining hydrodynamics and ML is proposed to predict water level dynamics as an indicator of compound flooding risk in a data-limited delta region. In this integrated approach, a hydrodynamic model first simulates various scenarios of compound flooding, and the outputs are then used to train the ML model.
A hybrid framework~\cite{fang2022hybrid} that integrates a hydrodynamic model, rapid flood model, and ML model is used for flood susceptibility assessments. 
Flood inventory data  is generated by the  hydrodynamic model. However, traditional methods in hydrodynamics for generating data typically involve running physical simulations or conducting experiments, which are often time-consuming. Hence, there is growing interest in generative ML approaches that can learn data distributions in unsupervised settings, potentially generating novel data beyond what traditional methods can produce. For example, FloodGAN~\cite{hofmann2021floodgan} utilizes an image-to-image translation approach where the model learns to generate 2D inundation predictions conditioned on heterogeneous rainfall distributions, employing a minimax game between two adversarial networks. 
Conditional GAN~\cite{farimani2017deep} can also be trained to simulate fluid flow based solely on observations, without knowledge of the underlying governing equations. However, a well-known drawback of GANs is their high sample complexity and lack of diversity. Therefore, several GANs based on PeML and PiML have been proposed to leverage prior knowledge of physics for hydrodynamic processes. For example, GAN-based models for simulating turbulent flows can be enhanced by incorporating conservation laws into the loss function~\cite{yang2019enforcing}.  Physics-informed GAN~\cite{meng2022learning} is employed to learn a functional prior from historical data. In addition, some
generative modeling approaches are developed to efficiently model data spaces with known invariances~\cite{shah2019encoding, joshi2020invnet}.  

However, most hydraulic problems use traditional hydrodynamic models to generate data. With the current development of generative models (such as GANs~\cite{creswell2018generative}, diffusion models~\cite{yang2022diffusion}, and generative pre-trained transformer~\cite{hadid2024geoscience,zhu2024foundations}) in the PaML community, the rapid generation of physical data based on generative models will gradually dominate.

\textbf{Uncertainty Quantification in Hydrodynamic Modeling.} See Fig.~\ref{fig:4} (label 2.7). Estimating uncertainty in hydrodynamic modeling is important for many applications. For instance, understanding, characterizing, and quantifying the impact of model and parametric uncertainties is crucial for making informed risk-based decisions in response to intense rainfall events.
Monte Carlo, first-order second-moment, and metamodeling are three popular methods to estimate the uncertainty of hydrodynamic models~\cite{dalledonne2019uncertainty}. Abbaszadeh et al.~\cite{abbaszadeh2022perspective} explore multiple origins of uncertainties across distinct layers of hydrometeorological and hydrodynamic model simulations as well as their complex interactions and cascading effects (e.g., uncertainty propagation) in forecasting flooding. Specifically, recent advances in PDgML techniques have
shown the benefits of using remote sensing data to correct elevation errors associated with building artifacts, flood defense structures, forests, and wetlands (such as object-based correction~\cite{cooper2019object}, FABDEM~\cite{hawker202230}, and random forest regression~\cite{gang2022flood}). In addition, Bayesian estimation/inversion methods are commonly employed to quantify and mitigate modeling uncertainties, such as employing a Markov chain Monte Carlo approach~\cite{siripatana2017assessing} or Bayesian calibration~\cite{schwindt2023bayesian}. Furthermore, PiML and PeML approaches have been gaining significant attention in the scientific community, as these methods can alleviate the computational burden and enhance the efficiency required for complex hydrodynamic modeling. For example, the physics-informed Gaussian process regression method~\cite{kohanpur2023urban} combined with the multi-level Monte Carlo for reducing the cost of estimating uncertainty  is designed for real-time flood depth forecasting. In addition, a deep probabilistic generative model~\cite{yang2019adversarial} implements a physics-based loss for uncertainty quantification, ensuring adherence to the structure imposed by PDEs. Bayesian PINNs~\cite{yang2021b} is proposed by using a Bayesian NN, which naturally encodes uncertainty. Bayesian PINNs~\cite{yang2021b} incorporate a PDE constraint to enforce the governing laws of the system as a prior for the Bayesian Network, enhancing prediction accuracy in scenarios with significant noise through physics-based regularization.  Another direction is a combination of physics knowledge and ML for uncertainty quantification. Hybrid frameworks that combine ML and hydrodynamic models can identify spatio-temporal features from historical flood events, aiding in predicting spatially distributed water depth and flood inundation extent caused by fluvial and coastal drivers~\cite{hosseiny2020framework}.

Integrating prior physics knowledge into ML for uncertainty quantification in hydrodynamic modeling has the potential to better characterize uncertainty. Further research in this area is recommended to fully leverage the benefits of the PaML community. PaML models have gained popularity in recent years as an efficient tool for hydrodynamic modeling, designing these emulators and their configurations remains challenging due to the uncertainties inherent in physical models, ML models, and their hybrid methods. There have been insufficient studies to rely on PaML and convincingly demonstrate its utility and effectiveness in solving challenging hydrodynamic problems.

\begin{figure}[!htp]
	\centering
	{\includegraphics[width = 0.85\textwidth]{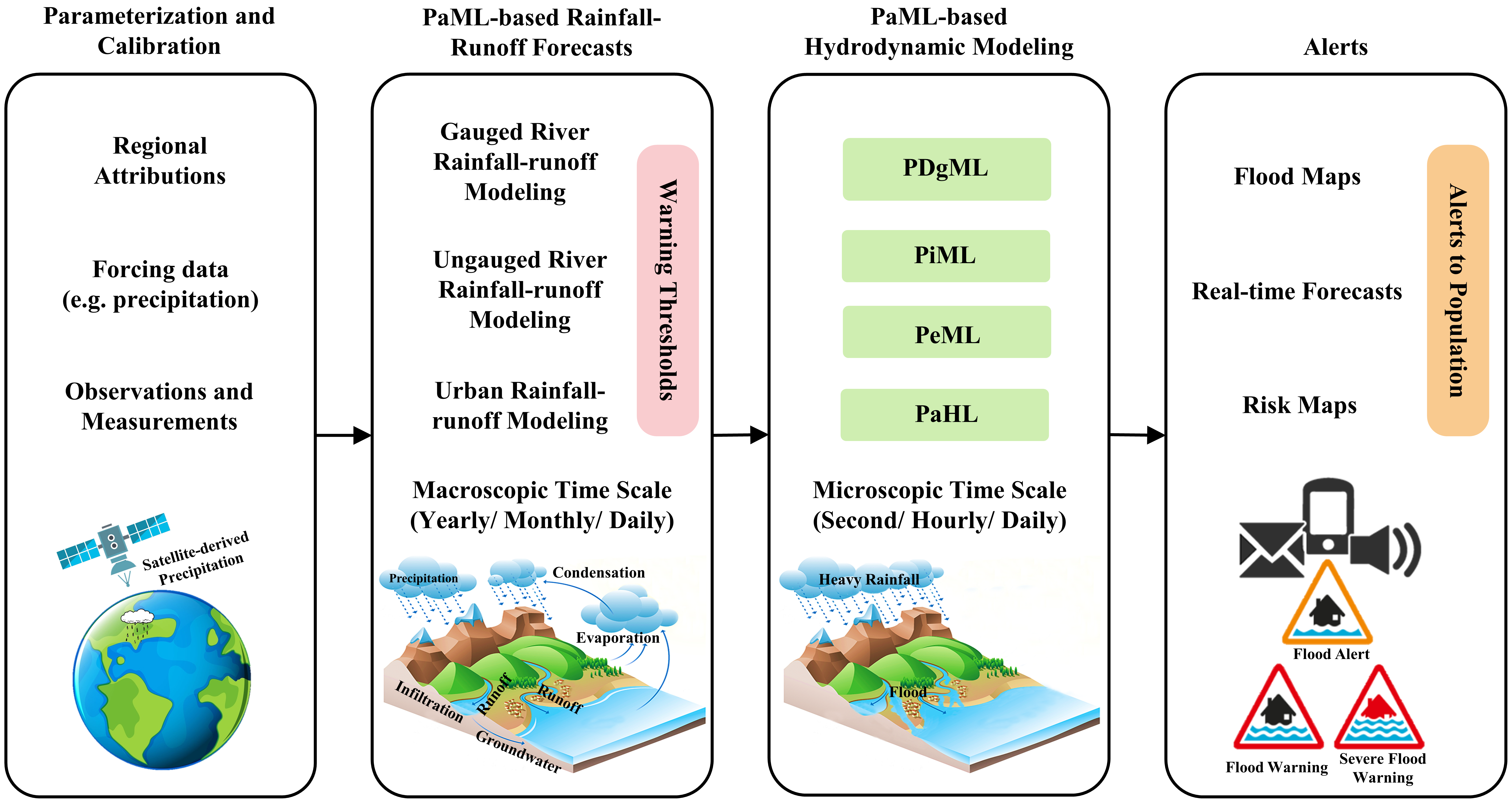}}
	\caption{The operational flood forecasting system based on HydroPML.}
	\label{fig:5}
\end{figure}

\begin{table}[!htp]
	\caption{\blue{Specific examples of HydroPML-based application highlights.}}
	\centering
	\resizebox{1.\textwidth}{!}
	{\Huge
\begin{tabular}{p{10cm}p{10cm}p{9cm}p{30cm}p{15cm}}
\hline
\textbf{Applications}                                     & \textbf{Examples}                                                                       & \textbf{Model Category}                 & \textbf{Model Details}                                                                                                                                                                                                                                                                                                                                                                                                                                                                                                                                                                                                                                                                                                                                                                                                                                                                                                                                                                                                                   & \textbf{Performances}                                                                                                                                                                                                                                              \\ \hline
Rainfall-induced landslide modeling and forecast & LandslideCast: A DL method for dynamic process modeling of real landslides~\cite{chen2024deep}     & PDgML (Deep Operator Networks) &\textbf{[Model]} The supervised FNO method is employed to learn the landslide dynamics process.\newline 
\textbf{[Input/framework/output]} The model inputs include topography, initial body height, base friction angle, and cohesion. The network maps these inputs to a high-dimensional representation space, iterates and updates them through four Fourier layers, and then maps the results to the output space. The model outputs the time series depth and speed, which reflect the dynamic characteristics of the landslide.                                                                                                                                                                                                                                                                                                                                                                                                                                             & The PDgML method replicates landslide dynamics with high performance, demonstrating excellent computational efficiency, generalization, and zero-shot downscaling capabilities.                                                                           \\ \hline
Flood modeling and forecast                      & FloodCast: Large-scale flood modeling and forecasting~\cite{xu2023large}                          & PiML (PINOs)                   & \textbf{[Model]} A geometry-adaptive physics-informed neural solver (GeoPINS) is introduced, which benefits from not requiring training data in PINNs and features a fast, accurate, and resolution-invariant architecture using FNO.  \newline \textbf{[Input/framework/output]} GeoPINS uses input channels composed of coordinates, time domain, and initial conditions of flood height, which are first lifted to a higher-dimensional representation by an MLP. Several Fourier layers then extract efficient spatiotemporal representations, and the outputs (flood depth) are obtained via MLP projection. GeoPINS is trained with a geometry-adaptive physics-constrained loss, which includes physics loss for controlling the residuals of 2-D depth-averaged shallow water equations and data loss for measuring the mismatch between the NN prediction and the initial/boundary conditions. & The experimental results for the 2022 Pakistan flood demonstrate that the GeoPINS enables high-precision, large-scale flood modeling. For instance, traditional hydrodynamics and GeoPINS exhibit exceptional agreement during high water levels. \\ \hline
Rainfall-runoff modeling and forecast            & MC-LSTM: Mass-conserving LSTM for rainfall-runoff modeling~\cite{hoedt2021mc, frame2023strictly}                     & PeML (PPeML)                   & \textbf{[Model]} MC-LSTM extends the inductive bias of the traditional LSTM to uphold conservation laws, ensuring mass input is conserved through modifications to the recurrent structure. \newline \textbf{[Input/framework/output]} The input of MC-LSTM includes precipitation, with auxiliary inputs comprising daily minimum and maximum temperature, solar radiation, vapor pressure, and 27 basin characteristics related to geology, vegetation, and climate. MC-LSTM directly learns the relationship between these inputs and streamflow or other output fluxes.                                                                                                                                                                                                                                                                                                                                                                                                       & MC-LSTM exhibits superior performance compared to other mass-conserving hydrological models, including the mesoscale hydrologic model (mHM), VIC, HBV, and others.                             \\ \hline
Hydrodynamic process understanding               & Physics-aware downsampling method: A DL method for scalable flood modeling~\cite{giladi2021physics,zhang2023pann} & PaHL (Neural–numerical)       & \textbf{[Model] }Physics-aware downsampling method for scalable 2D hydrodynamic modeling is proposed by minimizing the distance between the solutions on the fine and coarse grids. \newline \textbf{[Input/framework/output]} The model takes fine grid terrain as input and employs a downsampling neural network to transform it into coarse grid terrain. Subsequently, a parallel numerical solver with initial and boundary conditions on fine grid terrain is used to provide a supervised signal for the predictions on the coarse grids. This approach ensures that flood predictions on the coarse grids align closely with those on the fine grid.                                                                                                                                                                                                                                                                                                                                          & For hydrodynamic modeling, significant reductions in computational cost are achieved without compromising solution accuracy.                                                                                                                              \\ \hline
Rainfall-runoff-inundation modeling and forecast & The operational flood forecasting system based on HydroPML                     & Comprehensive                  & Four main modeling components include: (1) data management based on parameterization and calibration, (2) a rainfall-runoff forecasting model based on PaML, (3) a flood inundation model based on hydrodynamic models, and (4) alerts to the affected population.                                                                                                                                                                                                                                                                                                                                                                                                                                                                                                                                                                                                                                                                                                                                                              & Real-time flood forecast and flood level information are provided. The alerts and warnings comprise information about the forecasted water level, the inundation map, and the inundation depth.                                                           \\ \hline
\end{tabular}}
\label{tab10}
\end{table}
\subsection{Proposed HydroPML-based Application Highlights}

\blue{To illustrate the practical utility and performance of the HydroPML platform, we provide several specific examples of HydroPML-based application highlights, as shown in Table~\ref{tab10}. These include rainfall-induced landslide modeling and forecasting, flood modeling and forecasting, rainfall-runoff modeling and forecasting, hydrodynamic process understanding, and rainfall-runoff-inundation modeling and forecasting.
For rainfall-induced landslide modeling and forecasting, a supervised deep operator network (PDgML)~\cite{chen2024deep} is employed in HydroPML to learn the landslide dynamics process, replicating landslide dynamics with high performance. For flood modeling and forecasting, GeoPINS (PiML)~\cite{xu2023large} is used in HydroPML for large-scale flood forecasting. The experimental results for the 2022 Pakistan flood demonstrate that GeoPINS enables high-precision, large-scale flood modeling. For rainfall-runoff modeling and forecasting, MC-LSTM (PeML)~\cite{hoedt2021mc, frame2023strictly} is employed in HydroPML for rainfall-runoff modeling, exhibiting superior performance compared to other mass-conserving hydrological models. For hydrodynamic process understanding, the physics-aware downsampling method (PaHL)~\cite{giladi2021physics,zhang2023pann} is used in HydroPML for scalable 2D hydrodynamic modeling, achieving significant reductions in computational cost without compromising solution accuracy. The details of these models (such as category, input, framework, and output) are introduced in detail in Table~\ref{tab10}. The following presents a detailed application case of PaML's comprehensive methods: rainfall-runoff-inundation modeling and forecast.}

Based on the HydroPML technology, an operational flood forecasting (rainfall-runoff-inundation) system is developed. As shown in Fig.~\ref{fig:5}, 
four key modeling components are integrated into flood warning systems: (1) data management based on parameterization and calibration, (2) a rainfall-runoff forecasting model based on PaML, (3) a flood inundation model based on hydrodynamic models, and (4) alerts to the affected population.

The initial step involves the ingestion, calibration, and pre-processing of real-time data.  Specifically,   regional attributes (e.g., soil properties, LULC, and DEM) are acquired in near-real-time for specific  domains such as rivers or urban areas. 
Measurements and observations, such as water level station observation and water gauge measurements, are sourced from data providers or public organizations. Furthermore, precipitation data is obtained from the Integrated
Multi-satellite Retrievals for Global Precipitation
Measurement~\cite{hou2014global}
 or other reputable satellite-derived precipitation products. 
These data go through a series of calibration and parameterization processes: (1) unreasonable or negative manual errors in the data are corrected, and short periods of missing data are filled through linear interpolation; (2) these data are unified to the same spatio-temporal resolution through  resampling;
(3) these data are mapped to the input of rainfall-runoff forecasts through the parameterization module in HydroPML.  

In the second step, the PaML rainfall-runoff forecasting models  employ past precipitation data, historical measurements, and constant regional attributes to calculate predicted runoff for the specified target regions. 
The system encompasses three types of models: gauged river rainfall-runoff modeling based on short-term forecasts or long-term forecasts in HydroPML, ungauged river rainfall-runoff modeling utilizing  the missing data and ungauged basin module in HydroPML, and urban rainfall-runoff modeling based on  short-term forecasts or long-term forecasts for urban areas in HydroPML. After obtaining the future predicted runoff for a river or urban area, an evaluation of the potential flooding likelihood will be performed. Specifically, 
if the maximum forecasted runoff stage between the ``current time" of the forecast and the maximum lead time exceeds the predefined region-specific warning threshold, determined by local norms and historical water level data, this maximum stage is utilized to map flood inundation using PaML-based hydrodynamic modeling techniques.

To infer inundation extent and depth for a real-time flood forecast, hydrodynamic modeling based on PaML (such as PDgML, PiML, PeML, and PaHL) is 
used. Specifically, in the case of river floods, the forecasted discharge data obtained from rainfall-runoff modeling serves as the upstream boundary, while other model parameters (e.g., roughness) and downstream boundary conditions remain fixed as physical constants. In urban flood scenarios,  building footprints will serve as significant  physical boundary conditions. These physical boundaries for real-time flood simulation can be established through the implementation of optimization constraints (PiML), model framework constraints (PeML), or  neural–numerical hybrid learning techniques (PaHL). Upon receiving a real-time flood forecast, the simulation with the closest gauge water stage is identified, and the corresponding inundation and risk maps serve as the model output.

Finally, various flood levels, including flood alerts, flood warnings, and severe flood warnings, are disseminated to government authorities, emergency response agencies, and the affected population. These alerts and warnings include information about the forecasted water level, the inundation map, and the inundation depth.

\blue{Although these application cases primarily focus on the physical processes of water bodies, HydroPML can also be extended to other areas of hydrology, such as water quality. Recently, DL has become increasingly reliable for understanding and predicting water quality~\cite{wai2022applications,zhi2024deep}, such as water temperature and dissolved oxygen dynamics~\cite{zhi2023widespread,zhi2023temperature}. However, most approaches focus on water quality management and prediction under the PDgML frameworks. By integrating the physics of water-quality dynamics into ML frameworks (PeML), optimization processes (PiML), and hybrid learning (PaHL), the data requirements for ML can be reduced. This provides the possibility of achieving effective water-quality predictions for extreme events. Additionally, physic-discovery NNs can facilitate the discovery of new knowledge in water-quality dynamics~\cite{zhi2024deep}. Ultimately,  HydroPML offers a pathway for integrating data-driven ML methods and  physics-based models in hydrology. 
}



\section{Summary and Future Directions \label{section4}}
Emerging big data are advancing scientific research. The establishment of  Earth’s digital twin, particularly in the digital water cycle, relies on process-based hydrology and ML methodologies. However, process-based hydrology and ML are frequently regarded as separate paradigms in geosciences. Here, we introduce PaML as a transformative approach that bridges these paradigms, facilitating a paradigm shift in both fields.

PaML is a promising technique to take the best from both physics-based modeling and state-of-the-art ML models to improve the physical consistency, generalization, interpretability, and causality of ML. We present a comprehensive review of PaML methods, categorizing them into four aspects: physical data-guided ML, physics-informed ML, physics-embedded ML, and physics-aware hybrid learning. Advances in the PaML could promote ML-aided hypotheses in solving scientific problems, enabling not only rapidly exploiting the
information in big data. but also scientific discoveries derived from new theorizations.

We conduct a systematic review of process-based hydrology within the PaML community, specifically focusing on hydrodynamic processes and rainfall-runoff hydrological processes. Additionally, we release the HydroPML platform for the application of PaML to hydrological processes. Importantly, we highlight the most promising and challenging directions for different objectives and physics-aware ML methods. Specifically, from the hydrodynamic process perspective, we have four major recommendations, as follows.

\textbf{(1) PaML-based hydrodynamic solver}

 It is crucial to develop a comprehensive PaML-based hydrodynamic solver that spans different domains.
To rapidly solve hydrodynamic equations at various spatio-temporal scales and enhance the transferability and generalizability of models,  some  advanced PaML-based methods need to be  employed, along with the incorporation of both remote sensing data and in-situ measurements. 

\textbf{(2) Parameterization and calibration of hydrodynamic processes}

Physics-discovery NNs, data-physics-driven parameter discovery, or PaHL (such as differentiable parameter learning) need to be utilized for effective parameterization and calibration of hydrodynamic processes.

\textbf{(3) Data generation of hydrodynamic processes}

The use of generative models in PaML is necessary for creating virtual simulations of hydrodynamic processes under specific conditions. 

\textbf{(4) Uncertainty quantification in hydrodynamic modeling}

Integrating  prior physics knowledge into ML for uncertainty quantification in hydrodynamic modeling has the potential for improved characterization of uncertainty.

From the perspective of the rainfall-runoff process, we have identified three key recommendations, outlined as follows.

\textbf{(5) PaML-based short-term and long-term rainfall-runoff forecasts}

It is crucial  to explore PeML or hybrid models in an end-to-end manner for improved short-term and long-term forecasts. 

\textbf{(6) Spatio-temporal representation and reliability of rainfall-runoff forecasts}

Developing a robust spatio-temporal representation for different watersheds and enhancing the reliability of rainfall-runoff forecasts through the advanced PaML methods are ongoing research directions. 

\textbf{(7) Missing data, ungauged basins, parameterization, and uncertainty estimation in rainfall-runoff forecasts}

 Deeply leveraging the transferability of ML models and the interpretability of physical models will effectively address hydrological prediction challenges in ungauged basins, and integration of physics knowledge and ML techniques, such as PiML, PeML, and PaHL, enables efficient parameterization and better uncertainty estimation in rainfall-runoff forecasts. Furthermore, leveraging the PaML community and expanding hydrological data, we can seamlessly integrate pre-trained large models with domain-specific hydrological knowledge, establishing a foundational hydrology pre-training model for diverse rainfall-runoff prediction tasks.
 

 Overall, we suggest that future models  should be deeply integrated with physics-based models and data-driven models. This integration should be pursued through the creation of innovative frameworks and the facilitation of interdisciplinary and transdisciplinary collaborations.
  By doing so, we can not only enhance the explainability and causality of artificial general intelligence but also  lay the groundwork for the actualization of  Earth’s digital twin.
  
\section*{Acknowledgements}
The work is jointly supported by the German Federal Ministry of Education and Research (BMBF) in the framework of the international future AI lab "AI4EO -- Artificial Intelligence for Earth Observation: Reasoning, Uncertainties, Ethics and Beyond" (grant number: 01DD20001) and the project Inno$\_$MAUS (grant number: 02WEE1632B), by German Federal Ministry for Economic Affairs and Climate Action in the framework of the "national center of excellence ML4Earth" (grant number: 50EE2201C), by the Excellence Strategy of the Federal Government and the Länder through the TUM Innovation Network EarthCare and by Munich Center for Machine Learning.

\newpage
\section*{Supplementary Material}
\renewcommand{\tablename}{Supplementary Table}
\setcounter{table}{0}
\begin{table}[H]
	\caption{Proposed Physical Data-guided Machine Learning.}
	\centering
	\resizebox{1.05\textwidth}{!}
	{\Huge
	    \begin{tabular}{ccp{12cm}p{15cm}c}
		\hline
		\textbf{Categories}                                                       & \textbf{Types}                                                                                             & \textbf{Examples}                                                                                                             & \textbf{Objective(s)}                                                                                        & \textbf{Limitation(s)}                                                                                                                                   \\ \hline
		\multirow{11}{*}{\makecell{Classical Physical \\ Data-guided Neural Networks}} & \multirow{7}{*}{Spatio-temporal modeling}                                                         & LSTM~\cite{hochreiter1997long}                                                                 & Three gates to keep long-standing   relevant information and discard irrelevant information          & \multirow{6}{*}{Limited   generalization, high computational cost, limited physical interpretability}                                           \\
		&                                                                                                   & Physics-guided   LSTM~\cite{xie2021physics}                                                    & A deep learning network for   rainfall-runoff simulation                                            &                                                                                                                                                 \\
		&                                                                                                   & Informer~\cite{zhou2021informer}                                                               & Long sequence time-series   forecasting                                                             &                                                                                                                                                 \\
		&                                                                                                   & ConvLSTM~\cite{shi2015convolutional}                                                           & A spatio-temporal sequence   forecasting problem                                                     &                                                                                                                                                 \\
             &                                                                                                   & DisasterNets~\cite{xu2023disasternets}                                                           & Embedding machine learning in disaster mapping                                                     &                                                                                                                                                 \\
		&                                                                                                   & Earthformer~\cite{gao2022earthformer}                                                          & A space-time Transformer for Earth system forecasting                                               &                                                                                                                                                 \\
		&                                                                                                   & Geometric deep   learning~\cite{rozemberczki2021pytorch}                                       & A spatially-irregular forecast method                                                               &                                                                                                                                                 \\ \cline{2-5} 
		& \multirow{4}{*}{Dynamical system   modeling}                                                      & PDE-Net~\cite{long2018pde}                                                                     & A feed-forward deep   network-based PDE solver                                                      & Unable to handle temporally or   spatially sparse or non-uniform observations                                                                   \\
		&                                                                                                   & Mesh-Conv~\cite{hu2022mesh}                                                                    & A solver for modeling    non-uniform structured or unstructured spatial mesh                        & High computational cost                                                                                                                         \\
		&                                                                                                   & Neural message passing-based   models~\cite{sanchez2018graph, brandstetter2022message, wangmessage, li2018learning, ummenhofer2020lagrangian,sanchez2020learning}                                                                                & A general framework for supervised  learning on graphs (arbitrary domains)                          & Mesh-based or particle-based;   message-passing between small-scale moving and interacting objects                                              \\
		&                                                                                                   & DRL-based   method~\cite{ma2018fluid}                                                          & A coupled 2D control system                                                                         & Limited capabilities and   robustness; unknown behavior and convergence speed in action spaces of high   dimensionalities                       \\ \hline
		\multirow{8}{*}{\makecell{Deep Operator  \\ Networks}  }                      & \multirow{2}{*}{Deep Operator Networks   (DeepONet)~\cite{lu2021learning}}  & Multiple input DeepONet~\cite{jin2022mionet}                                                   & Employ DeepONet for realistic   setups (multiple initial conditions and boundary conditions)        & \multirow{2}{*}{Inaccurate   approximation for complex physical dynamics}                                                                       \\
		&                                                                                                   & HyperDeepONet~\cite{lee2022hyperdeeponet}                                                          & Enable the learning of the   DeepONet with a smaller set of parameters                              &                                                                                                                                                 \\ \cline{2-5} 
		& \multirow{2}{*}{Fourier neural   operator (FNOs)~\cite{li2020fourier}}      & Geo-FNO~\cite{li2022fourier}, Domain agnostic   FNO~\cite{liu2023domain} & Solve PDEs on arbitrary geometries                                                                  & \multirow{2}{*}{Unstable, limited   generalization, the basis functions in  the Fourier space are generally frequency localised   with no spatial resolution} \\
		&                                                                                                   & Factorized   FNO~\cite{tran2021factorized}                                                     & Improve  representation   stability for the operator                                                &                                                                                                                                                 \\ \cline{2-5} 
		& \multirow{2}{*}{Wavelet neural   operatos (WNOs)~\cite{tripura2022wavelet}} & WNO~\cite{tripura2022wavelet}                                                                  & Learn the network parameters in the wavelet space                                                & \multirow{2}{*}{Complexity,   limited  interpretability}                                                                                        \\
		&                                                                                                   & Coupled multiwavelets neural   operator~\cite{xiao2023coupled}                                                                              & Learn coupled PDEs in the wavelet space                                                          &                                                                                                                                                 \\ \cline{2-5} 
		& \multirow{2}{*}{Graph neural   operators (GNOs)~\cite{li2020neural}}        & Graph kernel networks~\cite{li2020neural}                                                      & An integral neural network that parameterizes Green’s   function                                 & \multirow{2}{*}{Difficult to train,   computational cost, limited generalization}                                                               \\
		&                                                                                                   & Nonlocal neural   operator~\cite{you2022nonlocal}                                              & A resolution-independent nonlocal kernel network                                                    &                                                                                                                                                 \\ \hline
		\multirow{5}{*}{\makecell{Physics-discovery   \\Neural Networks} }           & \multirow{4}{*}{Physics-discovery with unavailable prior knowledge}                               & Symbolic neural networks~\cite{sahoo2018learning,kim2020integration,sun2022symbolic}  & Infer concise equations without any prior knowledge                                             & \multirow{4}{*}{Challenging task due   to complex and nonlinear systems, as well as noisy and incomplete data}                                  \\
  &                                                                                                   & AI Feynman~\cite{udrescu2020ai,udrescu2020ai2}                                                              & Discover generalized symmetries in complex data                         &                                                                                                                                                 \\
		&                                                                                                   & GNN~\cite{cranmer2020discovering}                                                              & Identify the nontrivial relation   due to its excellent inductive capability                        &                                                                                                                                                 \\
		&                                                                                                   & Knowledge-agnostic   discovery~\cite{wang2021predicting, meempatta2019reviewing}               & Discover  the two-way feedback between human and   water systems                                    &                                                                                                                                                 \\ \cline{2-5} 
		& \multirow{2}{*}{Physics-discovery with   available prior knowledge}                               & SINDy~\cite{brunton2016discovering}                                                            & Discover governing equations   from data by sparse identification of nonlinear dynamical systems & \multirow{2}{*}{Limited high-fidelity   noiseless measurement}                                                                                  \\
		&                                                                                                   & PDE-FIND~\cite{rudy2017data}                                                                   & Data-driven discovery for PDEs                                                                      &                                                                                                                                                 \\ \hline
	\end{tabular}
}
	\label{1}
\end{table}

\begin{table}[!htp]
	\caption{Proposed Physics-informed Machine Learning.}
	\centering
	\resizebox{1.05\textwidth}{!}
	{\Huge
	\begin{tabular}{p{5cm}p{8cm}p{8cm}p{35cm}p{35cm}}
		\hline
		\textbf{Categories}                                            & \textbf{Types}                                                       & \textbf{Examples}                                                                                                 & \textbf{Objective(s)}                                                                                                                                                                                                                                                                                                                                                                                                                                                                                            & \textbf{Limitation(s)}                                                                                                                                                                                                                                                                                                                                                                                                                                                 \\ \hline
		& Vanilla PINNs                                               & PINNs~\cite{raissi2019physics}                                                     & A deep learning framework designed to address forward and inverse problems associated with nonlinear partial differential equations                                                                                                                                                                                                                                                                                                                                                                               & The challenging optimizations due to soft physics or PDE constraints; the challenge of nonlinear partial differential equations conveying information from initial or boundary conditions to unseen interior regions or future times; the sensitivity to hyperparameter selection; the slow training and re-optimization processes in practical applications and various instances; and the failure to converge on high-frequency, multi-scale, or time-dependent PDEs \\ \cline{2-5} 
		&        \multirow{9}{5cm}{Problem-specific \\ insights}                                                    & Universal PINNs~\cite{huang2022universal}                                          & Solve PDEs with a point source   by a lower bound constrained uncertainty weighting algorithm and a   multi-scale deep neural network                                                                                                                                                                                                                                                                                                                                                                 & Limited accuracy and generalization                                                                                                                                                                                                                                                                                                                                                                                                                           \\
		&                                                             & cPINN~\cite{jagtap2020conservative}                                                & 
Create a distinct conservative physics-informed neural network tailored for nonlinear conservation laws on discrete domains                                                                                                                                                                                                                                                                                                                                                                                   & Limited  representation   and parallelization capacity                                                                                                                                                                                                                                                                                                                                                                                                        \\
		&                                                             & nPINNs~\cite{pang2020npinns}                                                       & Extend PINNs to parameter and function   inference for integral      equations such as nonlocal Poisson and nonlocal turbulence models                                                                                                                                                                                                                                                                                                                        & Challenging optimization;   require more residual points                                                                                                                                                                                                                                                                                                                                                                                                      \\
		&                                                             & Improved   PINNs~\cite{pu2021solving}                                              & An improved PINN method for localized wave solutions of the derivative nonlinear Schrödinger equation in complex space, offering quicker convergence and optimal simulation performance                                                                                                                                                                                                                                                                                                              & Lack of consideration of complex integrable equations                                                                                                                                                                                                                                                                                                                                                                                                         \\
		&                                                             & AdaptivePINNs~\cite{wight2020solving}                                              & An automatic numerical solver   for the Allen-Cahn      and Cahn-Hilliard equations that utilizes an improved PINN                                                                                                                                                                                                                                                                                                                                          & Limited application range   (focused on the problem of solving differential equations)                                                                                                                                                                                                                                                                                                                                                                        \\
		&                                                             & CINN~\cite{braga2022characteristics}                                               & A simple  and efficient   characteristic-informed neural networks for solving forward and inverse problems in hyperbolic PDEs                                                                                                                                                                                                                                                                                                                                                                   & Limited application; limited   expression and generalization ability                                                                                                                                                                                                                                                                                                                                                                                          \\
		&                                                             & NSFnets~\cite{jin2021nsfnets}                                                      & Physics-informed neural networks   for the incompressible Navier-Stokes equations by automatic differentiation                                                                                                                                                                                                                                                                                                                                                                                          & Limited expressivity and   generalization; hyperparameters; failure to converge on high-frequency PDEs;   high computational cost                                                                                                                                                                                                                                                                                                                               \\
		&                                                             & GAPINN~\cite{oldenburg2022geometry}                                                & Geometry aware physics-informed   neural network for solving Navier–Stokes equation on irregular   geometries                                                                                                                                                                                                                                                                                                                                                                                       & Limited expressivity of   framework; cannot deal with time varying geometries                                                                                                                                                                                                                                                                                                                                                                                 \\
		&                & CPINNs~\cite{zhang2023cpinns}                                                      & Coupled physics-informed   neural networks designed for closed-loop geothermal systems                                                                                                                                                                                                                                                                                                                                                                                                                      & Limited expressivity of   framework; hyperparameters                                                                                                                                                                                                                                                                                                                                                                                                         \\ \cline{2-5} 
		&  \multirow{3}{5cm}{Input data}                                                           & Bayesian-PINNs~\cite{yang2021b}                                                    & Solve both forward and inverse   nonlinear problems described by PDEs and   noisy data                                                                                                                                                                                                                                                                                                                                                                                 & Limited  representation;   lack of large scale dataset testing                                                                                                                                                                                                                                                                                                                                                                                                \\
		&                                                             & ModalPINNs~\cite{raynaud2022modalpinn}                                             & An extension of physics-informed   neural networks incorporating enforced truncated Fourier for periodic flow reconstruction with a limited number of imperfect sensors                                                                                                                                                                                                                                                                                                                           & Limited application range                                                                                                                                                                                                                                                                                                                                                                                                                                     \\
		&                                 & $\Delta$-PINNs~\cite{costabal2022delta}                                                   & A positional encoding mechanism   for PINNs based on the eigenfunctions of the Laplace-Beltrami operator,   presented on complex geometries                                                                                                                                                                                                                                                                                                                                                           & Limited accuracy using finite elements; computing eigenfunctions of the Laplace-Beltrami operator   can become expensive for large meshes                                                                                                                                                                                                                                                                                                             \\ \cline{2-5} 
		&     \multirow{9}{5cm}{Frameworks}                                                         & PhyCRNet~\cite{ren2022phycrnet}                                                    & An encoder-decoder, ConvLSTM, is   proposed for low-dimensional spatial feature extraction and temporal  evolution learning. The loss function is   defined as the aggregated discretized PDE residuals                                                                                                                                                                                                                                                                                                   & Limited accuracy due to finite   difference                                                                                                                                                                                                                                                                                                                                                                                                                   \\
		&                                                             & PINNs with ensemble   models~\cite{haitsiukevich2022improved}                      & Train a set of ensemble PINNs and utilize the consensus (confidence) among the ensemble members as the criterion for extending the solution range into new domains                                                                                                                                                                                                                                                                                                                                                                & High computational cost                                                                                                                                                                                                                                                                                                                                                                                                                                       \\
		&                                                             & Physics-informed   GNNs~\cite{yan2022physics}                                      & Physics-informed graph neural   networks  not only provide a   representation that is specially tailored to the symmetries and properties of   the systems we consider, but they also are naturally invariant to the extent of   the system                                                                                                                                                                                                                                                                 & High computational cost; limited to homogeneous systems                                                                                                                                                                                                                                                                                                                                                                                                       \\
		&                                                             & MH-PINNs~\cite{zou2023hydra}                                                       & Multi-head physics-informed   neural network constructed as a potent tool for multi-task learning,   generative modeling, and few-shot learning for diverse problems                                                                                                                                                                                                                                                                                                                           & Limited expressivity of   framework; hyperparameters; difficult to optimize                                                                                                                                                                                                                                                                                                                                                                                  \\
		&                                                             & Phase2vec~\cite{ricci2022phase2vec}                                                & An embedding method that learns   high-quality, physically-meaningful representations of 2D dynamical systems by   a convolutional backbone                                                                                                                                                                                                                                                                                                                                                              & Failure in the n-dimensional case;   dynamics are qualitatively invariant to translations of the input coordinate   system                                                                                                                                                                                                                                                                                                                                           \\
		&                                                             & PI-PINN~\cite{tang2023physics}                                                     & Orthogonal polynomials are used   to construct the NN for solving nonlinear partial differential   equations                                                                                                                                                                                                                                                                                                                                                         & Limited expressivity of   framework                                                                                                                                                                                                                                                                                                                                                                                                                           \\
		&                                                             & PhyGeoNet~\cite{gao2021phygeonet}                                                  & Physics-constrained CNN learning   architecture to learn solutions of parametric PDEs on irregular domains   without any labeled data.                                                                                                                                                                                                                                                                                                                                                                  & Limited accuracy due to finite   difference; cannot solve dynamic geometrically irregular domains                                                                                                                                                                                                                                                                                                                                                             \\
		&                                                             & Spline-PINN~\cite{wandel2022spline}                                                & Training PDEs without any precomputed data using only a physics-informed loss function; provides rapid, continuous solutions that generalize across unseen domains                                                                                                                                                                                                                                                                                                             & Limited accuracy due to finite   difference                                                                                                                                                                                                                                                                                                                                                                                                                   \\
		&                                & Physics-constrained   Unsupervised Learning~\cite{michelis2022physics}             & Represent meshes naturally as   graphs, process them using graph networks, and formulate physics-based loss   to provide an unsupervised learning framework for PDEs                                                                                                                                                                                                                                                                                                                                    & Limited accuracy due to finite   difference                                                                                                                                                                                                                                                                                                                                                                                                                   \\ \cline{2-5} 
		&   \multirow{4}{5cm}{Improving speed and efficiency}                                                          & Meta-learning based PINNs                                                                                & de Avila Belbute-Peres et   al.~\cite{de2021hyperpinn} meta-trains a hyper-network that for each task can   generate weights for a small NN. Psaros et   al.~\cite{psaros2022meta} meta-learns a loss function that is used to   optimize the NN. Penwarden et al.~\cite{penwarden2021physics} proposes a   meta-learning approach that learns an initialization of weights such that the   NN can be optimized quickly. & Limited expressivity of   framework, unsteady in solving related PDEs                                                                                                                                                                                                                                                                                                                                                                                         \\
		&                                                             & XPINNs~\cite{jagtap2021extended}                                                   & A generalized space-time domain   decomposition framework, proposed for PINNs to solve nonlinear PDEs on   arbitrary complex-geometry domains                                                                                                                                                                                                                                                                                                                                                     & Hyperparameters; difficult to optimize                                                                                                                                                                                                                                                                                                                                                                                                                       \\
		&                                                             & canPINN~\cite{chiu2022can}                                                         & Coupled-automatic-numerical   differentiation framework that unifies the advantages of automatic differentiation   and numerical differentiation, providing more robust and efficient training   than automatic differentiation-based PINNs                                                                                                                                                                                                                                                                  & Limited application scenarios   (generally applicable to non-uniform samplings), limited computational   efficiency                                                                                                                                                                                                                                                                                                                                           \\
		&             & DT-PINNs~\cite{sharma2022accelerated}                                              & DT-PINNs are trained by   substituting precise spatial derivatives with high-order accurate numerical discretizations calculated using meshless radial basis function-finite differences and implemented via sparse matrix-vector multiplication                                                                                                                                                                                                                                                         & Limited accuracy due to radial   basis function-finite differences                                                                                                                                                                                                                                                                                                                                                                                            \\ \cline{2-5} 
		& Improving   generalization                                  & TPINN~\cite{manikkan2022transfer}                                                  & Transfer physics-informed   neural network for solving forward and inverse problems in nonlinear partial   differential equations by parameter sharing                                                                                                                                                                                                                                                                                                                                                  & Lack of consideration noise in input data                                                                                                                                                                                                                                                                                                                                                                                                                     \\ \cline{2-5} 
		&     \multirow{3}{5cm}{Multiscale and  multiphysics problems}                                                         & FBPINNs~\cite{dolean2022finite}                                                    & Finite basis physics-informed   neural networks use ideas from domain decomposition to accelerate the   learning process of PINNs and improve their accuracy                                                                                                                                                                                                                                                                                                                                             & Fail to coverge on   high-frequency PDEs; limited accuracy and expressivity                                                                                                                                                                                                                                                                                                                                                                                   \\
		&                                                             & Curriculum and  Sequence-to-sequence   PINN~\cite{krishnapriyan2021characterizing} & Curriculum regularization and   sequence-to-sequence learning, used to address the problems of PINNs that   fail to learn relevant physical phenomena for even slightly more   complex problems                                                                                                                                                                                                                                                                                               & Limited expressivity of   framework; hyperparameters; lack of realistic complex application                                                                                                                                                                                                                                                                                                                                                                  \\
		&    & Multi-scale   PINNs~\cite{wang2021eigenvector}                                     & Spatio-temporal and multi-scale   random Fourier features are used to address high-frequency or multi-scale   problems in PINN                                                                                                                                                                                                                                                                                                                                                                       & Hyperparameters                                                                                                                                                                                                                                                                                                                                                                                                                                              \\ \cline{2-5} 
		&  \multirow{2}{5cm}{Hyperparameter selection}                                                           & Self-adaptation PINNs~\cite{mcclenny2020self}                                      & Self-adaptation PINNs that make weights increase as the corresponding losses increase,   which is accomplished by training the network to simultaneously minimize   losses and maximize weights                                                                                                                                                                                                                                                                                 &   \multirow{2}{35cm}{Limited expressivity   of framework; the squared residual penalty method imposes a fundamental   limitation associated with conditioning}                                                                                                                                                                                                                                                                                                                                                                                                                                                              \\
		&               & Heuristic   method~\cite{van2022optimally}                                         & A heuristic method for   optimizing weighted loss functions for solving PDEs with NNs                                                                                                                                                                                                                                                                                                                                                                                                     &                                                                                                                                                                                                                                                                                                   \\ \cline{2-5} 
		&   \multirow{3}{5cm}{Data-physics-driven parameter discovery}                                                          & PINN-TFC~\cite{schiassi2021physics}                                                & Data-physics-driven parameters   discovery of problems modeled via ordinary      differential equations                                                                                                                                                                                                                                                                                                                                                & Cannot be applied to the data-driven discovery of problems                                                                                                                                                                                                                                                                                                                                                                                                        \\
		&                                                             & PINN-SR~\cite{saqlain2022discovering}                                                                                                  & Physics-informed neural network   with sparse regression to discover governing partial differential equations   from scarce and noisy data for nonlinear spatio-temporal systems                                                                                                                                                                                                                                                                                                                         & High computational cost; requiring a vast number of collocation points for complex systems; limitations of PINNs                                                                                                                                                                                                                                                                                                                                                 \\
		\multirow{-35}{5cm}{Physics-informed Neural Networks} &  & Discovering governing   equations~\cite{saqlain2022discovering}                    & Solve the inverse problem of   parameter identification in discrete, high-dimensional systems inspired   by physical applications                                                                                                                                                                                                                                                                                                                                                                  & Limitations of PINNs; cannot   address higher dimensional   settings                                                                                                                                                                                                                                                                                                                                             \\ \hline
		& DeepONet-based                                              & Physics-informed DeepONet~\cite{wang2021learning}                                  & Train the DeepONet by directly   incorporating known differential equations into the loss function, along with   some labeled datasets                                                                                                                                                                                                                                                                                                                                                                 & Limitations on fixed grid or   basis; cannot achieve discretization invariant; limitation of linear   approximation; limited computational efficiency and generalization                                                                                                                                                                                                                                                                                      \\ \cline{2-5} 
		&           \multirow{3}{5cm}{FNO-based}                                                   & PINO~\cite{li2021physics}                                                          & An integration of Fourier   operator learning and physics-informed settings                                                                                                                                                                                                                                                                                                                                                                                                                             & Limited accuracy on solving   non-periodic problems; limited adaptability due to the selection of model   parameters; it cannot be applied to large-scale and long-term sequence PDE   solution                                                                                                                                                                                                                                                               \\
		&                                                             & FC-PINO~\cite{maust2022fourier}                                                    & An architecture that leverages   Fourier continuation is proposed to apply the exact gradient method to PINO   for nonperiodic problems                                                                                                                                                                                                                                                                                                                                                                 & Limited adaptability due to the   selection of model parameters; it cannot be applied to large-scale and   long-term sequence PDE solution; it cannot achieve geometry-adaptive learning                                                                                                                                                                                                                                                                      \\
		&                                 & GeoPINS~\cite{xu2023large}                                                         & A geometry-adaptive   physics-informed neural solver based on the advantages of no   training data in physics-informed neural networks, as well as possessing a   fast, accurate, geometry-adaptive and resolution-invariant architecture                                                                                                                                                                                                                                                   & Limited adaptability due to the   selection of model parameters                                                                                                                                                                                                                                                                                                                                                                                               \\ \cline{2-5} 
		\multirow{-5}{5cm}{\makecell[tl]{Physics-\\informed \\ Neural \\ Operators}}   & WNO-based                                                   & Physics-informed WNO~\cite{tripura2023physics}                                     & A physics-informed WNO for learning solution operators of families of parametric PDEs without labeled training data, leveraging the time-frequency localization advantage of wavelets                                                                                                                                                                                                                                                                                           & Complexity; the absence of a   standardized procedure to obtain the optimum and desired network architecture   of a physics-informed WNO and to assign the weights of boundary/initial   condition loss; the lack of information regarding how to learn the operator   for systems involving multi-scale physics systems                                                                                                                                      \\ \hline
	\end{tabular}  }
\label{2}
\end{table}

\begin{table}[!htp]
	\caption{Proposed Physical Equation-embedded Machine Learning.}
	\centering
	\resizebox{1.05\textwidth}{!}
	{\Huge
	\begin{tabular}{p{12cm}p{8cm}p{25cm}p{25cm}}
		\hline
		\textbf{Methods}                                                                         & \textbf{Physical equations}                                           & \textbf{Embedding methods and   objectives}                                                                                                                                                                                                                                                                                                                                                                                                                                                                                                                                                                    & \textbf{Limitation(s)}                                                                                                                         \\ \hline
		AutoIP~\cite{long2022autoip}                              & Differential equations                                       & Automatically   incorporating physics that can integrate all kinds of differential equations   into Gaussian processes                                                                                                                                                                                                                                                                                                                                                                                                                                                                               & Limitations on   large-scale applications                                                                                             \\
		FINN~\cite{karlbauer2022composing}                        & Advection-diffusion type   equations                         & Learning spatio-temporal   advection-diffusion processes by modeling the constituents of PDEs in a compositional manner (the incorporation   of explicitly defined physical equations)                                                                                                                                                                                                                                                                                                                                                                             & Limited scalability   (beyond second-order spatial derivatives); limitation on the real-world   applications                         \\
		DeLaN~\cite{lutter2019deep}                               & Euler-Lagrange equation   from Lagrangian mechanics          & Deep Lagrangian networks   encoding the Euler-Lagrange equation originating, which can be trained using   standard end-to-end optimization techniques while maintaining physical   plausibility                                                                                                                                                                                                                                                                                                                                                                                                      & Limited representational power of   the physics prior when applied to a physical system with complex dynamics  \\
		MeshGraphNets~\cite{pfaff2020learning}                    & Mesh-based partial   differential equations                  & A framework for learning   mesh-based simulations using graph neural networks: the mesh-space, spanned   by the simulation mesh, and the Euclidean world-space in which the simulation   manifold is embedded                                                                                                                                                                                                                                                                                                                                                                                         & High computational   complexity; hyperparameters                                                                                      \\
		FGN~\cite{li2022graph}                                    & A Lagrangian viewpoint   of the Navier-Stokes equation       & Fluid graph networks   (FGN) uses graphs to represent the fluid field. In FGN, fluid particles are   represented as nodes and their interactions are represented as edges. FGN   decomposes the simulation scheme into separate parts, including advection,   collision, and pressure projection                                                                                                                                                                                                                                                                                                      & Limited generalization   and computational efficiency                                                                                 \\
		Continuous   convolutions~\cite{ummenhofer2020lagrangian} & Lagrangian fluid   simulation of the Navier-Stokes equation & An efficient ConvNet   architecture, based on continuous convolution layer, processes sets of   particles. Dynamic particles and static particles are used to represent the   fluid and describe the boundary of the scene, respectively                                                                                                                                                                                                                                                                                                                                                                & Limited computational   efficiency; cannot deal with deformable solids                                                                \\
		Differentiable   modules                                                        & Differential equations                                       & These modules implement physical   equations as differentiable feature space or computational graphs, enabling   the optimization of dynamic processes with analytical gradients and therefore   improving sample efficiency. Examples include differentiable physics-informed graph   networks~\cite{seo2019differentiable}, a differentiable layer   (PDE-Constrained-Layer)~\cite{negiar2022learning}, physics-aware difference   graph networks~\cite{seo2020physics}, physics-based finite difference convolution connection~\cite{rao2023encoding}, and
  differentiable   modeling~\cite{shen2023differentiable} & Limitations on   large-scale applications; require an in-depth understanding of differential   equations                             \\ \hline
	\end{tabular}}
\label{3}
\end{table}

\subsection*{Appendix A. Physical Methods in Hydrological Process}

The hydrological model can be categorized into three main groups: empirical models, conceptual models, and physically-based models~\cite{devia2015review}. Empirical models utilize existing datasets without accounting for the specific features and processes of hydrological systems, and are therefore referred to as data-driven models. On the other hand, physical and conceptual models require a comprehensive understanding of the movement of surface water in the hydrological cycle~\cite{srinivasulu2008rainfall}.

Conceptual models describe runoff processes by linking simplified components of the hydrological cycle using interconnected storages to represent different components. These models are typically lumped and employ uniform parameter values across the entire watershed. They need a variety of parameters and meteorological input data. Calibration of these models requires a substantial amount of meteorological and hydrological records. A  host of conceptual models have been proposed in the past including HBV model~\cite{bergstrom1992hbv}, non-recorded catchment areas (NRECA)~\cite{normad1981hydrologic}, Boughton model~\cite{boughton1984simple}, ARNO
model~\cite{todini1996arno}, Xinanjiang model~\cite{singh1995computer}, etc.

Physical-based models are grounded in an understanding of the physics governing hydrological processes. These models utilize fundamental physical laws and principles, including equations for water balance, conservation of mass and energy, momentum, and kinematics. Typically, finite difference or finite computation schemes are employed to solve these equations. For being built on physical laws and equations, spatially-explicit, process-based models are, in theory, capable to respond correctly to dynamic changes in catchment conditions and climate forcing; they can thus be used for the assessment of non-linearities between states, drivers, and catchment responses~\cite{brunner2021extremeness,basso2023extreme}. Therefore, they are practical tools to predict, project, and attribute impacts of global change (e.g. climate change, land use change, water management) on the water cycle and even hydrological extremes.  
The primary advantage of physical models lies in their ability to connect model parameters with physical characteristics of the catchment, thereby enhancing their realism. Since physical models are semi- or fully distributed across watersheds, they need a significant amount of data such as soil-hydraulic properties, vegetation, land cover characteristics, initial water depth, topography, and dimensions of the river network. This data is crucial for investigating changes in the hydrological cycle resulting from human activities and climate change, which are vital for effective water resource management~\cite{leavesley1994modeling}.  Some examples of physical models include soil and water assessment tool (SWAT)~\cite{arnold1998large}, 
system hydrologique european (SHE)~\cite{abbott1986introduction} or MIKE SHE~\cite{refsgaard1995mike}, visualizing ecosystem land management assessments (VELMA)~\cite{mckane2014visualizing}, kinematic runoff and erosion model (KINEROS)~\cite{singh1995computer}, Penn State integrated hydrologic modeling system (PIHM)~\cite{qu2007semidiscrete}, variable
infiltration capacity (VIC)~\cite{wood1992land}, and others. 

\subsection*{Appendix B. Physical Methods in Hydrodynamic Modeling}

Depending on the spatial representation of the flow, hydrodynamic models can be dimensionally grouped into one-dimensional (1D)~\cite{caleffi2003finite}, two-dimensional (2D)~\cite{doherty2011use}, and three-dimensional (3D) models~\cite{prakash2014modelling}. 


\textbf{1D models.} 
1D hydrodynamic models have been widely used in modeling water flows~\cite{lin2006integrating}. Many hydraulic situations can make the 1D assumption, either because a more detailed solution is unnecessary (e.g., knowledge in other dimensions is not required for the purpose) or because the flow is distinctly 1D, such as in a channelized flow or a confined channel.  Typically, 1D models solve equations derived to ensure conservation of mass and momentum, resulting in the well-known 1D Saint-Venant equations,
\begin{equation}
\begin{gathered}
\frac{\partial h}{\partial t}+u \frac{\partial h}{\partial x}=q, \\
\frac{\partial u}{\partial t}+u \frac{\partial u}{\partial x}+g \frac{\partial h}{\partial x}+g\left(S_f-S\right)=0,
\end{gathered}
\end{equation}
where $x$ denotes the distance along the river channel; $t$ denotes time; $q$ represents the water inflow per unit length of the channel from land surface and subsurface runoff, groundwater, and precipitation; $u$ and $h$ represent the dynamics of velocity and water depth along the river channel, respectively; $g$ represents gravity; $S$ represents the riverbed slope; and $S_f$ represents the friction slope, which can be computed using the Chezy–Manning equation~\cite{bjerklie2005comparison}.

\textbf{2D models.}
The 2D models depict water or flood flow as a two-dimensional field with the assumption that the third dimension, water depth, is shallow compared to the other two dimensions~\cite{de2013applicability,de2012improving,xu2023large}. Most methods solve two-dimensional shallow water equations, which describe conservation of mass and momentum in a plane and are derived by depth-averaging the Navier-Stokes equations,
\begin{equation}
\begin{gathered}
\frac{\partial h}{\partial t}+\frac{\partial q_x}{\partial x}+\frac{\partial q_y}{\partial y}=0,\\
\underbrace{\frac{\partial q_x}{\partial t}}_{\substack{\text { local } \\
		\text { acceleration }}}+\underbrace{\frac{\partial}{\partial x}\left(u q_x\right)+\frac{\partial}{\partial y}\left(v q_x\right)}_{\begin{array}{c}
	\text { convective } \\
	\text { acceleration }
	\end{array}}+\underbrace{g h \frac{\partial(h+z)}{\partial x}}_{\begin{array}{c}
	\text { pressure }+ \\
	\text { bed gradients }
	\end{array}}+\underbrace{\frac{g h^2\|\mathbf{q}\| q_x}{h^{7 / 3}}}_{\text {friction }}=0,\\
\underbrace{\frac{\partial q_y}{\partial t}}_{\begin{array}{c}
	\text { local } \\
	\text { acceleration }
	\end{array}}+\underbrace{\frac{\partial}{\partial y}\left(v q_y\right)+\frac{\partial}{\partial x}\left(u q_y\right)}_{\begin{array}{c}
	\text { convective } \\
	\text { acceleration }
	\end{array}}+\underbrace{g h \frac{\partial(h+z)}{\partial y}}_{\begin{array}{c}
	\text { pressure }+ \\
	\text { bed gradients }
	\end{array}}+\underbrace{\frac{g n^2\|\mathbf{q}\| q_y}{h^{7 / 3}}}_{\text {friction }}=0, 
\end{gathered}
\end{equation}
where the two Cartesian directions are represented by $x$ and $y$; $q_x$ and $q_y$ denote the $x$ and $y$ components of the discharge per unit width vector $\mathbf{q}$; $u$ and $v$ are the $x$ and $y$ components of the flow velocity, respectively; $z$ denotes the bed elevation; and $n$ represents the Manning’s friction coefficient. In flood situations, the significant simplification introduced by the local inertial approximation is based on the assumption that the convective acceleration terms are negligible compared to other terms, allowing them to be disregarded.

\textbf{3D models.} 
For various water flow scenarios, there has been a tendency to consider the detailed representation of flow dynamics in 3D~\cite{teng2017flood}. For example, it is crucial to model vertical turbulence, vortices, and floods caused by dam breaks. Consequently, 3D models have been developed to incorporate vertical features. Some of these models solve horizontal flow using 2D shallow water equations and incorporate a quasi-three-dimensional extension to represent velocity in vertical layers. Other 3D models are based on the 3D Navier-Stokes equations, which describe the movement of fluid and are typically formulated as,
\begin{equation}
\begin{gathered}
\frac{\partial u}{\partial t}+u \cdot \nabla u+\frac{1}{\rho} \nabla p=g+\mu \nabla \cdot \nabla u, \\
\nabla \cdot u=0,
\end{gathered}
\end{equation}
where $u$ represents velocity; $\rho$ denotes fluid density; $p$ stands for pressure; $g$ is gravitational acceleration; and $\mu$ represents kinematic viscosity. The incompressibility condition ($\nabla \cdot u = 0$) assumes that material density remains constant within a fluid parcel. Based on the process representation, these models can be categorized into two primary types: grid-based Eulerian models and particle-based Lagrangian models.

\clearpage
\newpage

\bibliographystyle{naturemag-doi} 
\bibliography{sample}

\end{document}